# Temporal Decision Trees:
# Model-based Diagnosis of Dynamic Systems On-Board

**Luca Console**                                                LUCA.CONSOLE@DI.UNITO.IT
**Claudia Picardi**                                            CLAUDIA.PICARDI@DI.UNITO.IT
*Dipartimento di Informatica, Università di Torino,*
*Corso Svizzera 185, I-10149, Torino, Italy*

**Daniele Theseider Dupré**                                       DTD@MFN.UNIPMN.IT
*Dipartimento di Informatica, Università del Piemonte Orientale*
*Spalto Marengo 33, I-15100, Alessandria, Italy*

## Abstract

The automatic generation of decision trees based on off-line reasoning on models of a domain is a reasonable compromise between the advantages of using a model-based approach in technical domains and the constraints imposed by embedded applications. In this paper we extend the approach to deal with temporal information. We introduce a notion of temporal decision tree, which is designed to make use of relevant information as long as it is acquired, and we present an algorithm for compiling such trees from a model-based reasoning system.

## 1. Introduction

The embedding of software components inside physical systems became widespread in the last decades due to the convenience of including electronic control into the systems themselves. This phenomenon occurs in several industrial sectors, ranging from large-scale products such as cars to much more expensive systems like aircraft and spacecrafts.

The case of automotive systems is paradigmatic. In fact, the number and complexity of vehicle subsystems which are managed by software control increased significantly since the mid 80s and will further increase in the next decades (see Foresight-Vehicle, 2002), due to the possibility of introducing, at costs that are acceptable for such wide scale products, more flexibility in the systems, for e.g. increased performance and safety, and reduced emissions. Systems such as fuel injection control, ABS (to prevent blockage of the wheels while braking), ASR (to avoid slipping wheels), ESP (controlling the stability of the vehicle), would not be possible at feasible costs without electronic control.

The software modules are usually installed on dedicated Electronic Control Units (ECUs) and they play a very important role since they have complete control of a subsystem: human "control" becomes simply an input to the control system, together with inputs from appropriate sensors. For example, the position of the accelerator pedal is an input to the ECU which controls fuel delivery to the injectors.

A serious problem with these systems is that the software must behave properly also in presence of faults and must guarantee high levels of availability and safety for the controlled system and for the vehicle. The controlled systems, in fact, are in many cases safety critical: the braking system is an obvious example. This means that monitoring the systems behaviour, detecting and isolating failures, performing the appropriate recovery actions is a critical task that must be performed by control software. If any problem is detected or suspected the software must react, modifying the way the system is controlled, with the primary goal of guaranteeing safety and availability. According to recent estimates, about 75% of the ECU software deals with detecting problems and performing recovery actions, that is to the tasks of diagnosis and repair (see again Foresight-Vehicle, 2002).

Thus the design of the diagnostic software is a very critical and time consuming activity, which is currently performed manually by expert engineers who use their knowledge to perform the "Failure Mode and Effect Analysis (FMEA)" [1] and define diagnostic and recovery strategies.

The problem is complex and critical per-se, but it is made even more difficult by a number of other issues and constraints that have to be taken into account:

- The resources that are available on-board must be limited, in terms of memory and computing power, to keep costs low. This has to be combined with the problem that near real time performance is needed, especially in situations that may be safety critical. For example, for direct injection fuel delivery systems, where fuel is maintaned at a very high pressure (more than 1000 bar) there are cases where the system must react to problems within a rotation of the engine (e.g. 15 milliseconds at 4000 rpm), to prevent serious damage of the engine and danger to passengers. In fact, a fuel leakage can be very dangerous if it comes from a high pressure line. In this case it is important to distinguish whether a loss of pressure is due to such a leak, in order to activate some emergency action (for example, stop the engine), or to some other failure which can simply be signalled to the user.

- In order to keep costs acceptable for a large scale product, the set of sensors available on board is usually limited to those necessary for controlling the systems under their correct behaviour; thus, it is not always easy to figure out the impact that faults may have on the quantities monitored by the sensors, whose physical, logical and temporal relation to faults may be not straightforward.

- The devices to be diagnosed are complex from the behavioural point of view: they have a dynamic and time-varying behaviour; they are embedded in complex systems and they interact with other subsystems; in some cases the control system automatically compensates deviations from the nominal behaviour.

These aspects make the design of software modules for control and diagnosis very challenging but also very expensive and time consuming. There is then a significant need for improving this activity, making it more reliable, complete and efficient through the use of automated systems to support and complement the experience of engineers, in order to meet the growing standards which are required for monitoring, diagnosis and repair strategies.

Model-based reasoning (MBR) proved to be an interesting opportunity for automotive applications and indeed some applications to real systems have been experimented in the 90s (e.g., Cascio & Sanseverino, 1997; Mosterman, Biswas, & Manders, 1998; Sachenbacher, Malik, & Struss, 1998; Sachenbacher, Struss, & Weber, 2000). The type of models adopted in MBR are conceptually not too far away from those adopted by engineers. In particular, the component oriented approach typical of MBR fits quite well with the problem of dealing with several variants of systems, assembled starting from the same set of basic components. For a more thorough discussion of the advantages of the MBR approach, see Console and Dressler (1999).

---

1. The result of FMEA is a table which lists, for each possible fault of each component of a system, the effect of the faults on the component and on the system as a whole and the possible strategy to detect the faults. This table is compiled manually by engineers, based on experience knowledge and on a blueprint of the system.

Most of the applications developed so far, however, concentrated on off-board diagnosis, that is diagnosis in the workshop, and not on on-board diagnosis. The case of on-board systems seems to be more complicated since, due to the restrictions on hardware to be put on-board, it is still questionable if diagnostic systems can be designed to reason on first-principle models on-board. For this reason other approaches have been developed in order to exploit the advantages of MBR also in on-board applications. In particular, a *compilation-based* scheme to the design of on-board diagnostic systems for vehicles was experimented in the Vehicle Model Based Diagnosis (VMBD) BRITE-Euram Project (1996-99), and applied to a Common-rail fuel injection system (Cascio, Console, Guagliumi, Osella, Sottano, & Theseider, 1999). In this approach a model-based diagnostic system is used to generate a compact on-board diagnostic system in the form of a decision tree. Similarly, automated FMEA reports generated by Model-Based Reasoning in the Autosteve system can be used to generate diagnostic decision trees (Price, 1999). Yet similar is the idea proposed by Darwiche (1999), where diagnostic rules are generated from a model in order to meet resource constraints.

These approaches have interesting advantages. On the one hand, they share most of the benefits of model-based systems, such as relying on a comprehensive representation of the system behaviour and a well defined characterization of diagnosis. On the other hand, decision trees and other compact representations make sense for representing on-board diagnostic strategies, being efficient in space and time. Furthermore, algorithms for synthesizing decision trees from examples are well established in the machine learning community. In this specific case the examples are the solutions (diagnoses and recovery actions) generated by a model-based system.

However, the basic notion of decision tree and the approaches for learning such trees from examples have a major limitation for our kind of applications: they do not cope properly with the temporal behaviour of the systems to be diagnosed, and, in particular, with the fact that incremental discrimination of possible faults, leading to a final decision on an action to be taken on-board, should be based on observations acquired *across time*, thus taking into account temporal patterns.

For such a reason, in the work described in this paper we introduce a new notion of decision tree, the *temporal decision tree*, which takes into account the temporal dimension, and we introduce an algorithm for synthesizing temporal decision trees.

Temporal decision trees extend traditional decision trees in the fact that nodes have a temporal label which specifies when a condition should be checked in order to select one of the branches or to make a decision. As we shall see, this allows taking into account that in some cases the order and the delay between observable measures influences the decision to be made and thus provides important power to improve the decision process. Waiting, however, is not always possible and thus the generation of the trees includes a notion of deadline for each possible decision. Thus, the temporal decision process supports the possibility of selecting the best decision, exploiting observations and their temporal locations (patterns) and taking into account that in some cases at some point a decision has to be taken anyway to prevent serious problems.

The rest of the paper is organized as follows. In section 2 we summarize some basic ideas about model-based diagnosis (MBD), the use of decision trees in conjunction with it, and the temporal dimension in MBD and in decision trees. In section 3 we provide basic formal definitions about decision trees, which form the basis for their extension to temporal decision trees in section 4. We then describe (section 5) the problem of synthesizing temporal decision trees and our solution (section 6).

## 2. Model-based Diagnosis and Decision Trees

In this section we briefly recall the basic notions of model-based diagnosis and we discuss how decision trees can be used for diagnostic purposes, focusing on how they have been used in VMBD in conjunction with the model-based approach (Cascio et al., 1999).

### 2.1 The Atemporal Case

First of all let us sketch the atemporal case and the "traditional" use of diagnostic decision trees.

#### 2.1.1 Atemporal model-based diagnosis

The starting point for model-based diagnosis is a model of the structure and behaviour of the device to be diagnosed. More specifically, we assume a component centered approach in which:

- A model is provided for each component type; a component is characterized by

    - A set of variables (with a distinguished set of interface variables);
    - A set of *modes*, including an *ok* mode (correct behaviour) and possibly a set of fault modes.
    - A set of relations involving component variables and modes, describing the behaviour of the component in such a mode. These relations may model the correct behaviour of the device and, in some cases, the behaviour in presence of faults (faulty behaviour).

- A model for the device is given as a list of the component instances and of their connections (connections between interface variables).

In the Artificial Intelligence approach, models are usually qualitative, that is the domain of each variable is a finite set of values. Such an abstraction has proven to be useful for diagnostic purposes.

The model can be used for simulating the behaviour of a system and then for computing diagnoses. In fact, given a set of observations about the system behaviour, diagnoses can be determined after comparing the behaviour predicted by the model (in normal conditions or in the presence of single or multiple faults) and the observed behaviour.

In order for the model to be useful for on-board diagnosis, for each fault mode $F$ (or combination of fault modes) the model must include the recovery action the control software should perform in case $F$ occurs. In general these actions have a cost, mainly related to the resulting reduction of functionality of the system. Moreover, two actions $a_1$ and $a_2$ can be related in the following sense:

- $a_1$, the recovery action associated with fault $F_1$, carries out operations that *include* those performed by $a_2$, the recovery action associated with $F_2$;

- $a_1$ can be used as a recovery action also when $F_2$ occurs; it may however carry out unneeded operations, thus reducing the system functionality more than strictly necessary.

- However, in case we cannot discriminate between $F_1$ and $F_2$, applying $a_1$ is a rational choice.

In section 4.3 we will present a model for actions which formalizes this relation.

Thus the main goal of the on-board diagnostic procedure is to decide the best action to be performed, given the observed malfunction. This type of procedure can be efficiently represented using decision trees.

### 2.1.2 Decision trees

Decision trees can be used to implement classification problem solving and thus some form of diagnostic procedure. Each node of the tree corresponds to an observable. In on-board diagnosis, observables correspond either to direct sensor readings, or to the results of computations carried out by the ECUs on the available measurements. In the following the word "sensor" will denote both types of observables; it is worth pointing out that the latter may require some time to be performed. In this paper we mainly assume that a sensor reading takes no time; however the approach we propose deals also with the case in which a sensor reading is time consuming, as pointed out in section 5.1. A node can have as many descendants as the number of qualitative values associated with the sensor. The leaves of the tree correspond to actions that can be performed on board. Thus, given the available sensor readings, the tree can be very easily used to make a decision on the recovery action to be performed.

Such decision trees can be generated automatically from a set of examples or cases. By example here we mean a possible assignment of values to observables and the corresponding diagnosis, or possible alternative diagnoses, and a selected recovery action which is appropriate for such a set of suspect diagnoses. This set can be produced using a model-based diagnostic systems, which, given a set of observables can compute the diagnoses and recovery actions.

In the atemporal case, with finite qualitative domains, the number of possible combinations of observations is finite, and usually small, therefore considering all cases exhaustively (and not just a sample) is feasible and there are two equivalent ways of building such an exhaustive set of cases:

1. *Simulation approach:* for each fault $F$, we run the model-based system to predict the observations corresponding to $F$.

2. *Diagnostic approach:* we run a diagnosis engine on combinations (all relevant combinations) of observations, to compute the candidate diagnoses for each one of these cases.

In either case, the resulting decision tree contains the same information as the set of cases; if, once sensors are placed in the system, observations have no further cost, the decision tree is just a way to save space with respect to a table, and speed up lookup of information.

In this way the advantages of the model-based approach and of the use of compact decision trees can be combined: the model-based engine produces diagnoses based on reusable component models and can be used as the diagnoser off-board; compact decision trees, synthesized from cases classified by the model-based engine, can be installed on-board.

## 2.2 Towards Temporal Decision Trees

In this section we briefly recall some basic notions on temporal model-based diagnosis (see Brusoni, Console, Terenziani, & Theseider Dupré, 1998 for a general discussion on temporal diagnosis), and we informally introduce temporal decision trees.

### 2.2.1 Temporal MBD

The basic definition of MBD is conceptually similar to the atemporal case. Let us consider the main differences.

As regards the model of each component type we consider a further type of variable: state variables used to model the dynamic behaviour of the component. The set of relations describing the behavior of the component (for each mode) is augmented with temporal information (constraints); we do not make specific assumptions on the model of time, even though, as we shall see in the following, this has an impact on the cases which can be considered for the tree generation. As an example, these constraints may specify a delay between an input and an output or in the change of state of the component.

As regards recovery actions, a deadline for performing the action must be specified; this represents the maximum time that can elapse between fault detection and the recovery action; this is the amount of time available for the control software to perform discrimination. This piece of information is specific to each component instance, rather than component type, because the action and the deadline are related to the potential unacceptable effects that a fault could have on the overall system; the same fault of the same component type could be very dangerous for one instance and tolerable for another.

Diagnosis is started when observations indicate that the system is not behaving correctly. Observations correspond to (possibly qualitative) values of variables across time. In general, in the temporal case a diagnosis is an assignment of a mode of behaviour to component instances across time such that the observed behaviour is explained by the assignment given the model. For details on different ways of defining explanation in this case see Brusoni et al. (1998). For the purposes of this paper we are only interested in the fact that, given a set of observables, a diagnosis (or a set of candidate diagnoses if no complete discrimination is possible) can be computed and a recovery action is determined.

This means that the starting point of our approach is a table containing the results of running the model-based diagnostic system on a set of cases, (almost) independently of the model-based diagnostic system used for generating the table.

We already mentioned that in the static case, with finite qualitative domains, an exhaustive set of cases can be considered. In the temporal case, if the model of time is purely qualitative, a table with temporal information cannot be built by prediction, while it can be built running the diagnosis engine on a set of cases with quantitative information: diagnoses which make qualitative predictions that are inconsistent with the quantitative information can be ruled out. Of course, this cannot in general be done exhaustively, even if observations are assumed to be acquired at discrete times; if it is not, the decision tree generation will actually be *learning* from examples.

Thus a simulation approach can only be used in case the temporal constraints in the model are precise enough to predict at least partial information on the temporal location of the observations, e.g., in case the model includes quantitative temporal constraints.

The diagnostic approach can be used also in case of weaker (qualitative) temporal constraints in the model.

As regards the observations, we consider the general case where a set of snapshots is available; each snapshot is labelled with the time of observation and reports the value of the sensors (observables) at that time. This makes the approach suitable for different notions of time in the underlying model and on the observations (see again the discussion in Brusoni et al., 1998).

**Example 1** *The starting point for generating a temporal decision tree is a table like the one in Figure 1.*

| | $s_1$ | | | | | | | | $s_2$ | | | | | | | | $s_3$ | | | | | | | | **Act** | **Dl** |
|---|---|---|---|---|---|---|---|---|---|---|---|---|---|---|---|---|---|---|---|---|---|---|---|---|---|---|
| | 0 | 1 | 2 | 3 | 4 | 5 | 6 | 7 | 0 | 1 | 2 | 3 | 4 | 5 | 6 | 7 | 0 | 1 | 2 | 3 | 4 | 5 | 6 | 7 | | |
| **sit₁** | n | n | n | n | h | h | | | l | v | v | v | v | v | | | n | n | n | l | l | l | | | a | 5 |
| **sit₂** | h | h | h | | | | | | h | n | n | | | | | | n | n | n | | | | | | b | 2 |
| **sit₃** | n | n | n | n | h | h | h | h | l | l | l | l | l | v | v | v | v | n | n | n | l | l | l | v | v | b | 7 |
| **sit₄** | n | n | n | n | h | h | h | h | l | l | l | l | l | l | l | v | n | n | h | h | h | h | h | h | c | 6 |
| **sit₅** | h | h | h | h | | | | | h | n | n | n | | | | | n | n | n | l | | | | | c | 3 |
| **sit₆** | n | n | n | h | h | h | | | l | v | v | z | z | z | | | n | n | n | l | l | v | | | d | 5 |
| **sit₇** | h | h | h | h | h | h | | | l | l | l | n | n | l | v | | n | n | n | l | l | v | | | b | 5 |
| **sit₈** | h | h | h | h | h | h | | | h | h | n | n | l | l | l | | n | n | n | l | v | z | | | c | 5 |

Figure 1: An example of a set of cases for learning temporal decision trees.

*Each row of the table corresponds to a situation (case or "example" in the terminology of machine learning) and it reports:*

- *For each sensor $s_i$ the values that have been observed at each snapshots (in the example we have 8 snapshots, labelled as 0 to 7);* **n**, **l**, **h** *and* **v** *correspond to the qualitative values of the sensor measurements;* **n** *for* normal, **l** *for* low, **h** *for* high, **v** *for* very low *and* **z** *for* zero.

- *The recovery action* **Act** *to be performed in that situation.*

- *The deadline* **Dl** *for performing such an action.* ◇

A table as the one in the above example represents a set of situations that may be encountered in case of faults and, as noticed above, it can be generated using either a diagnostic or a simulation approach. In the next section we shall introduce the notion of temporal decision trees and show how the pieces of information about sensor histories like those in the table above can be exploited in the generation of such trees.

### 2.2.2 INTRODUCTION TO TEMPORAL DECISION TREES

Traditional decision trees do not include a notion of time, i.e., the fact that data may be observable at different times or that different faults may be distinguished only by the temporal patterns of data. Thus, neglecting the notion of time may lead to limitations in the decision process.

For such a reason in this work we introduce a notion of temporal decision tree. Let us analyse the intuition behind temporal decision trees and the decision process they support. Formal definitions will be provided later in the paper.

Let us consider, for example, the fault situations $\mathbf{sit}_3$ and $\mathbf{sit}_4$ in Figure 1, and let us assume, for the sake of simplicity, that the only available sensor is $s_2$. The two fault situations have to be distinguished in the control software because they require different recovery actions. Both of them can be detected by the fact that $s_2$ shows a **l**ow value. Moreover, in both situation after a while $s_2$ starts showing a **v**ery low value.

The only way to discriminate these two situations is to make use of temporal information, that is to exploit the fact that in $\mathbf{sit}_3$ value **v** shows up after 4 time units from fault detection, while in $\mathbf{sit}_4$ the same value shows up after 6 time units.

In order to take into account this in a decision tree, we have to include time into the tree. In both examples, the best decision procedure is to wait after observing thst $s_2 = \mathbf{l}$ (that is, after dectecting that a fault has occurred). After 4 time units we can make a decision, depending on whether $s_2 = \mathbf{v}$ or not. This corresponds to the procedure described by the tree in Figure 2.

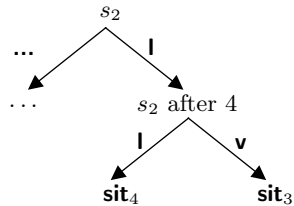

Figure 2: A simple example of temporal decision tree

Obviously, waiting is not always the solution or is not always possible. In many cases, in fact, safety or other constraints may impose some deadlines for performing recovery actions. This has to be reflected in the generation of the decision procedure. Suppose, in the example above, that the deadline for $\mathbf{sit}_3$ was 3 rather than 6: in this case the two situations would have been indistinguishable, beacause it would have been infeasible to wait 4 time units.

Thus, an essential idea for generating small decision trees in the temporal case is to take advantage of the fact that *in some cases there is nothing better than waiting*[2], in order to get a good discrimination, provided that safety and integrity of the physical system are kept into account, and that deadlines for recovery actions are always met.

More generally, one should exploit all the information about temporal patterns of observables and about deadline, like the ones in Figure 1, to produce an optimal diagnostic procedure.

Another idea we use in our approach is the integration of incremental discrimination, which is the basis for the generation and traversal of a decision tree, with the incremental acquisition of information across time.

In the atemporal case, the decision tree should be generated in order to guide the incremental acquisition of information: different subtrees of a node are relative to different sets of faults and therefore may involve different measurements: in a subtree T we perform the measurements that are useful for discriminating the faults compatible with the measurements that made us reach subtree T, starting from the root. For off-board diagnosis, this allows reducing the average number of measurements to get to a decision (i.e. the average depth of the tree), which is useful because measurements have a cost - e.g. time for an operator to take them from the system; for on-board diagnosis, even in case all measurements are simply sensor readings, which have no cost once the sensor has been made part of the system, we are interested in generating *small* decision trees to save space.

In the temporal case there is a further issue: the incremental acquisition of information is naturally constrained by the flow of time. If we do not want to store sensor values across time — which seems a natural choice since we have memory constraints — information must be acquired when it is available and it is not possible to read sensors once the choice of waiting has been made. This issue will be taken into account in the generation of temporal decision trees.

## 3. Basic Notions on Decision Trees

Before moving to a formal definition of temporal decision trees, in this section we briefly recall some definitions and algorithms for the atemporal case. In particular, we recall the standard **ID3** algorithm (Quinlan, 1986), which will be the basis for our algorithm for the temporal case. The definitions in this section are the standard ones (see any textbook on Artificial Intelligence, e.g., Russel & Norvig, 1995).

### 3.1 Decision Trees

We adopt the following formal definition of decision tree, which will be extended in section 4.1 to temporal decision trees.

**Definition 1** Let us consider a decision process $\mathcal{P}$ where $\mathbb{A}$ is the set of available decisions, $\mathbb{O}$ is the set of tests that can be performed on the external environment, and $\mathbf{out}(o_i) =$ $\{v_1, \ldots, v_{k_i}\}$ are the possible outcomes of test $o_i \in \mathbb{O}$. A *decision tree* for $\mathcal{P}$ is a labelled tree structure $\mathbf{T} = \langle r, N, E, \mathcal{L} \rangle$ where:

- $\langle r, N, E \rangle$ is a tree structure with root $r$, set of nodes $N$ and set of edges $E \subseteq N \times N$; $N$ is partitioned in a set of *internal* nodes $N_I$ and a set of *leaves* $N_L$.

- $\mathcal{L}$ is a labelling function defined over $N \cup E$.

- If $n \in N_I$, $\mathcal{L}(n) \in \mathbb{O}$; in other words each internal node is labelled with the name of a test.

- If $(n, c) \in E$ then $\mathcal{L}((n, c)) \in \mathbf{out}(\mathcal{L}(n))$; that is, an edge directed from $n$ to $c$ is labelled with a possible outcome of the test associated with $n$.

- Moreover, if $(n, c_1), (n, c_2) \in E$ and $\mathcal{L}((n, c_1)) = \mathcal{L}((n, c_2))$ then $c_1 = c_2$, and for each $v \in \mathbf{out}(\mathcal{L}(n))$ there is $c$ such that

---

2. A different approach would be that of weighing the amount of elapsed time agains the possibility of better discriminating faults; such an approach is something we are considering in the future work on temporal decision trees, as outlined in section 7.

  $(n, c) \in E$ and $\mathcal{L}((n, c)) = v$; that is, $n$ has exactly one outgoing edge for each possible outcome of test $\mathcal{L}(n)$.

- If $l \in N_L$, $\mathcal{L}(l) \in \mathbb{A}$; in other words each leaf is labelled with a decision. $\diamond$

When the decision-making agent uses the tree, it starts from the root. Every time it reaches an inner node $n$, the agent performs test $\mathcal{L}(n)$, observes its outcome $v$ and follows the $v$-labelled edge. When the agent reaches a leaf $l$, it makes decision $a = \mathcal{L}(l)$.

## 3.2 Building Decision Trees

Figure 3 shows a generic recursive algorithm that can be used to build a decision tree starting from a set of `Examples` and a set of `Tests`.

Recursion ends when either the remaining examples do not need further discrimination because they all correspond to the same decision, or all available observables have been used, and the values observed match cases with different decisions. In the latter case observables are not enough for getting to the proper decision and if an agent is actually using the tree, it should take into account this.

In case no terminating condition holds, the algorithm chooses an observable variable `test` to become the root label for subtree $\mathbf{T}$. Depending on how CHOOSETEST is implemented we get different specific algorithms and different decision trees.

A subtree is built for each possible outcome `value` of `test` in a recursive call of BUILDTREE, with sets `Tests_Update` and `SubExamples` as inputs. `Tests_Update` is obtained by removing `test` from the set of tests, in order to avoid using it again. `SubExamples` is the subset of `Examples` containing only those examples that have `value` as outcome for `test`.

As mentioned before, there are as many specific algorithms, and, in general, results, as there are implementations of CHOOSETEST. It is in general desirable to generate a tree with minimum average depth, for two reasons:

- Minimizing average depth means minimizing the average number of tests and thus speeding up the decision process.

- In machine learning, a small number of tests also means a higher degree of generalization on the particular examples used in building the tree.

```
1    function BuildTree (set Examples, set Tests)
2    returns a decision tree T = ⟨root, Nodes, Edges, Labels⟩
3    begin
4        if all ex ∈ Examples correspond to the same decision then
5            return BuildLeaf(Examples);
6        if Tests is empty then
7            return BuildLeaf(Examples);
8        test ← ChooseTest (Tests, Examples);
9        root ← new node;
10       Nodes ← {root}; Edges ← ∅; Labels(root) ← test;
11       T ← ⟨root, Nodes, Edges, Labels⟩;
12       Tests_Update ← Tests \ {test};
13       for each possible outcome value of test do begin
14           SubExamples ← {ex ∈ Examples | test has outcome value in ex};
15           if SubExamples is not empty then begin
16               SubTree ← BuildTree (SubExamples, Tests_Update);
17               Append(T, root, SubTree);
18               Labels((root, Root(SubTree))) ← value;
19           end;
20       end;
21       return T;
22   end.
```

Figure 3: Generic algorithm for building a decision tree

### 3.3 ID3

Unfortunately, finding a decision tree with minimum average depth is an intractable problem; however, there exists a good best-first heuristic for choosing tests in order to produce trees that are "not too deep". This heuristic was proposed in the **ID3** algorithm (Quinlan, 1986), and is base on the concept of *entropy* from *information theory*. In the following we recall this approach in some detail also in order to introduce some notation which will be used in the rest of the paper.

**Definition 2** Given a (discrete) probability distribution $\mathbf{P} = \{p_1, \ldots, p_n\}$ its *entropy* $\mathcal{E}(\mathbf{P})$ is:

$$(1) \qquad \mathcal{E}(\mathbf{P}) = -\sum_{i=1}^{n} p_i \log_2 p_i$$

$\diamond$

Entropy measures the *degree of disorder*. When we choose a test, we want it to split the examples with the lowest degree of disorder with respect to the decisions associated with them.

Given a set of examples $\mathbb{E}$ we introduce the sets:

$$\mathbb{E}\,|_a = \{e \in \mathbb{E} \mid \text{the decision associated with example } e \text{ is } a\}.$$

If the set of available decisions is $\mathbb{A} = \{a_1, \ldots, a_n\}$ then $\mathbb{E}\,|_{\mathbb{A}} = \{\mathbb{E}\,|_{a_1}, \ldots, \mathbb{E}\,|_{a_n}\}$ is a partition of $\mathbb{E}$.

**Definition 3** For each $a_i \in \mathbb{A}$, $i = 1, \ldots, n$, we define the *probability of $a_i$ with respect to* $\mathbb{E}$ as follows:

$$P(a_i; \mathbb{E}) = \frac{|(\mathbb{E}|_{a_i})|}{|\mathbb{E}|}$$

$\diamond$

It is worth noting that, if examples are endowed with their *a priori* probabilities, we can redefine $P(a_i; \mathbb{E})$ in order to take them into account. The basic formulation of **ID3** assumes however that all examples are equiprobable and it computes the probability distribution from the frequencies of the examples.

The entropy of $\mathbb{E}$ is:

(2)
$$\mathcal{E}(\mathbb{E}) = -\sum_{i=1}^{n} P(a_i; \mathbb{E}) \log_2 P(a_i; \mathbb{E}).$$

If all decisions are equiprobable, we get $\mathcal{E}(\mathbb{E}) = \log_2 n$, which is the maximum degree of disorder for $n$ decisions. If all decisions but one have probability equal to 0, then $\mathcal{E}(\mathbb{E}) = 0$: the degree of disorder is minimum.

Entropy is used as follows for test selection. A test $o$ with possible outcomes $v_1, \ldots, v_k$, splits the set of examples into:

$$\mathbb{E}|_{o \rightarrow v_i} = \{e \in \mathbb{E} \mid \text{test } o \text{ has value } v_i \text{ in } e\}.$$

$\mathbb{E}|_o = \{\mathbb{E}|_{o \rightarrow v_1}, \ldots, \mathbb{E}|_{o \rightarrow v_k}\}$ is again a partition of $\mathbb{E}$. In particular, if while building the tree we choose test $o$, $\mathbb{E}|_{o \rightarrow v_j}$ is the subset of examples we use in building the subtree for the child corresponding to $v_j$. The lowest the degree of disorder in $\mathbb{E}|_{o \rightarrow v_j}$, the closer we are to a leaf. Therefore, following equation (2):

$$\mathcal{E}(\mathbb{E}|_{o \rightarrow v_j}) = -\sum_{i=1}^{n} P(a_i; \mathbb{E}|_{o \rightarrow v_j}) \log_2 P(a_i; \mathbb{E}|_{o \rightarrow v_i}).$$

Finally, we define the entropy of a test $o$ as the average entropy on its possible outcomes:

**Definition 4** The *entropy of a test $o$ with respect to a set of examples* $\mathbb{E}$ is:

(3)
$$\mathcal{E}(o; \mathbb{E}) = \sum_{j=1}^{k} P(o \rightarrow v_j) \mathcal{E}(\mathbb{E}|_{o \rightarrow v_j}),$$

where[3] $P(o \rightarrow v_j) = \dfrac{|(\mathbb{E}|_{o \rightarrow v_j})|}{|\mathbb{E}|}$.

$\diamond$

The **ID3** algorithm simply consists of choosing the test with lowest entropy. Figure 4 shows the implementation of CHOOSETEST that yields **ID3**.

---

3. Again, if examples are endowed with *a priori* probabilities, this definition can be changed in order to take them into account

```
1    function ID3ChooseTest (set Tests, set Examples)
2    returns a test best_test
3    begin
4        best_test ← any element in Tests;
5        min_entropy ← log₂ |Examples|;
6        for each test ∈ Tests do begin
7            part ← Partition({Examples}, test);
8            ent ← Entropy(part);
9            if (ent < min_entropy) then begin
10               min_entropy ← ent;
11               best_test ← test;
12           end;
13       end;
14       return best_test;
15   end.
```

Figure 4: **ID3** implementation of ChooseTest

## 4. Extending Decision Trees

In this section we formally introduce the notion of temporal decision tree, and we show how timing information can be added to the set of examples used in tree building. Moreover we introduce a model for recovery action that expresses information needed by the algorithm.

### 4.1 Temporal Decision Trees

In section 2.2.2 we motivated the monotonicity requirement for temporal decision trees, so that their traversal requires information relative to increasing time and then no information has to be stored.

We now discuss how temporal information is actually included in the tree and matched with temporal information on the observations. The tree has to be used when some abnormal value is detected for some sensor (fault detection). We then intend the time of fault detection as the *reference time* for temporal labels of observations in the tree. If we look for example at the data shown in Figure 1, we see that, for every fault situation, there is always at least one sensor whose value at time 0 is different than **n**normal. The reason is that, for each fault situation, we associate a 0 time label to the first snapshot in which there is a sensor showing some deviation from nominal behaviour.

The following definition provides the extension for the temporal dimension in decision trees.

**Definition 5** A *temporal decision tree* is a decision tree $\langle r, N_I, N_L, E, \mathcal{L} \rangle$ endowed with a time-labelling function $\mathcal{T}$ such that:

**(1)** $\mathcal{T} : N_I \mapsto \mathbb{R}^+$; we call $\mathcal{T}(n)$ a *time label*;

**(2)** if $n' \in N_I$ and there exist $n$ such that $(n, n') \in E$ (in other words, $n'$ is child of $n$), then $\mathcal{T}(n') \geq \mathcal{T}(n)$. ◇

Since we assume not to store any information, but rather to use information for traversing the tree as dictated by the tree itself, a first branching for discrimination is provided depending on *which* is the sensor that provided such a value. We then assume to have different temporal decision trees, one for each sensor which could possibly provide the first abnormal value, or, alternatively, that the root node has no time label, and the edges from the root are not labelled with different values of a single observable, but with different sensors which could provide the first abnormal observation. Each tree, or subtree in the second alternative, can be generated independently of the other ones, only using the examples where the sensor providing fault detection is the same. This generation is what will be described in the rest of the paper.

Let us tree how a temporal decision tree (or forest, in the case of multiple detecting sensors) can be exploited by an on-board diagnostic agent in order to choose a recovery action. When the first abnormal value is detected, the agent activates a time counter and starts visiting the appropriate tree from the root. When it reaches an inner node $n$, the agent retrieves both the associated test $s = \mathcal{L}(n)$ and the time label $t = \mathcal{T}(n)$. Then it waits until the time counter reaches $t$, performs test $s$ and chooses one of the child nodes depending on the outcome. When the agent reaches a leaf, it performs the corresponding recovery action.

With respect to the atemporal case, the agent has now the option to wait. From the point of view of the agent it may seem pointless to wait when it could look at sensor values, since reading sensor values has no cost. However, from the point of view of the tree things are quite different: we do not want to add a test that makes the tree deeper and at the same time is not necessary.

Condition **(2)** states that the agent can only move forward in time. This corresponds to the assumption that sensor readings are not stored, discussed in section 2.2.2.

**Example 2** *Let us consider the diagnostic setting described in example 1. Figure 5 shows a temporal decision tree for such setting, that is a temporal decision tree that uses the sensors and recovery actions mentioned in Figure 1. If such a tree is run on the fault situations in the table, a proper recovery action is selected within the deadline.* ⋄

## 4.2 Adding Timing Information to the Set of Examples

In order to generate temporal decision trees, we need temporal information in the examples. We already introduced informally the notion of a set of examples (or "fault situations") with temporal information when describing the table in Figure 1. The following definition formalizes the same notion.

**Definition 6** A *temporal example-set (te-set for short)* $\mathbb{E}$ is a collection of *fault situations* $\mathbf{sit}_1, \dots, \mathbf{sit}_n$ characterized by a number of sensors $\mathbf{sens}_1, \dots, \mathbf{sens}_m$ and an ascending sequence of time labels $t_1 < \dots < t_{\mathbf{last}}$ representing the instants in time for which sensor readings are available. In this context we call *observation* a pair $\langle \mathbf{sens}_i, t_j \rangle$. A te-set is organized in a table as follows:

**(1)** The table has $n$ rows, one for each fault situation.

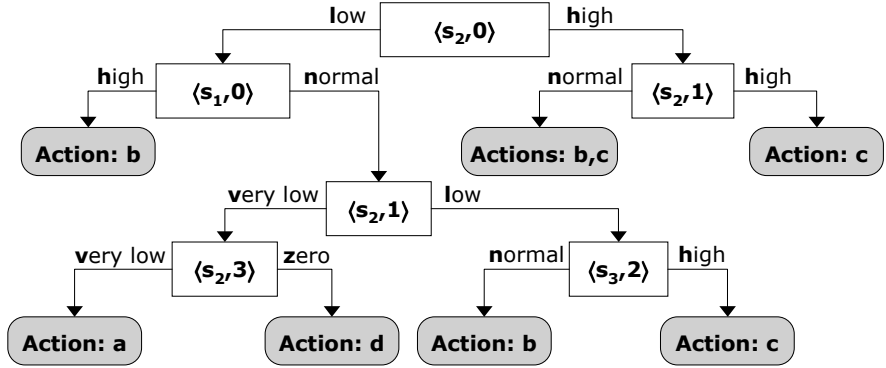

Figure 5: A temporal decision tree for the situations described in Figure 1.

**(2)** The table has $m \times$ **last** observation columns containing the outcomes of the different observations for each fault situation. We denote by **Val**$(\mathbf{sit}_h, \langle \mathbf{sens}_i, t_j \rangle)$ the value measured by sensor $\mathbf{sens}_i$ at time $t_j$ in fault situation $\mathbf{sit}_h$.

**(3)** The table has a distinguished column **Act** containing the recovery action associated with each fault situation. We denote by **Act**$(\mathbf{sit}_h)$ the recovery action associated with $\mathbf{sit}_h$.

**(4)** The table has a second distinguished column **Dl** containing the deadline for each fault situation. We denote by **Dl**$(\mathbf{sit}_h)$ the deadline for $\mathbf{sit}_h$, and we have that **Dl**$(\mathbf{sit}_h) \in \{t_1, \dots, t_{\mathbf{last}}\}$ for each $h = 1, \dots, n$. We define a *global deadline* for a te-set $S$ as **Dl**$(S) = \min\{\mathbf{Dl}(\mathbf{sit}) \mid \mathbf{sit} \in S\}$.

We moreover assume that a probability[4] $P(\mathbf{sit}; \mathbb{E})$ is associated with each $\mathbf{sit} \in \mathbb{E}$, such that $\sum_{\mathbf{sit} \in \mathbb{E}} P(\mathbf{sit}; \mathbb{E}) = 1$. For every $\mathbb{E}' \subset \mathbb{E}$ and for every $\mathbf{sit} \in \mathbb{E}'$ we introduce the following notation: $P(\mathbb{E}'; \mathbb{E}) = \sum_{\mathbf{sit} \in \mathbb{E}'} P(\mathbf{sit}; \mathbb{E})$ and $P(\mathbf{sit}; \mathbb{E}') = \frac{P(\mathbf{sit}; \mathbb{E})}{P(\mathbb{E}'; \mathbb{E})}$. $\diamond$

### 4.3 A Model for Recovery Actions

The algorithms we shall introduce require a more detailed model of recovery actions. In particular we want to better characterize what happens when it is not possible to uniquely identify *the* most appropriate recovery action. Moreover, we want to quantify the loss we incur in when this happens.

We start with a formal definition:

**Definition 7** A *basic model for recovery actions* is a triple $\langle \mathbb{A}, \prec, \chi \rangle$ where:

**(1)** $\mathbb{A} = \{a_1, \dots, a_K\}$ is a finite set of symbols denoting *basic recovery actions*.

---

4. $P(\mathbf{sit}; \mathbb{E})$ can be computed as a frequency, that is $P(\mathbf{sit}; \mathbb{E}) = 1/n$, where $n$ is the number of fault situations, or it can be known *a priori* and added to the set of examples.

**(2)** $\prec\, \subseteq\, \mathbb{A}\times\mathbb{A}$ is a partial strict order relation on $\mathbb{A}$. We say that $a_i$ is *weaker* than $a_j$, written as $a_i \prec a_j$, if $a_j$ produces more recovery effects than $a_i$, in the sense that $a_j$ could be used in place of $a_i$ (but not the vice versa). We therefore assume that there are no drawbacks in actions, that is any action can be performed at any time with no negative consequences, apart from the cost of the action itself (see below). This is clearly a limitation and something to be tackled in future work (see the discussion in section 7).

**(3)** $\chi : \mathbb{A} \mapsto \mathrm{I\!R}^+$ is the *cost* function, and is such that if $a_i \prec a_j$ then $\chi(a_i) < \chi(a_j)$. $\chi$ associates a cost with each basic recovery action, expressing possible drawbacks of the action itself. Recovery actions performed on-board usually imply a performance limitation or the abortion of some ongoing activity; costs are meant to estimate monetary losses or inconveniences for the users resulting from these. The requirement of monotonicity with respect to $\prec$ stems from the following consideration: if $a_i \prec a_j$ and $\chi(a_i) \geq \chi(a_j)$ it would not make any sense to ever perform $a_i$, since $a_j$ could be performed with the same effects at the same (or lower) cost. We moreover assume that costs are independent from the fault situation (a consequence of the no-drawbacks assumption mentioned in the previous point). $\diamond$

**Example 3** *Let us consider again the four recovery actions $a, b, c, d$ that appear in the te-set of Figure 1. Figure 6 shows a basic action model for them. The graph expresses the oredering relation $\prec$, while costs are shown next to action names.* $\diamond$

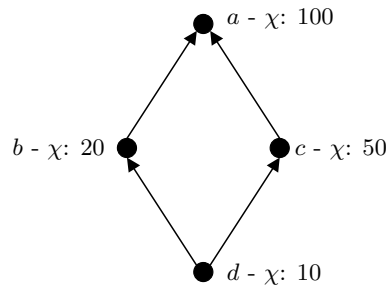

Figure 6: A basic action model.

We have seen in the previous section that with each fault situation is associated a recovery action; usually this association depends on the fault, but it may also depend on the operating conditions in which the fault occurs.

What happens when we cannot discriminate multiple fault situations? In section 3.2, while outlining the generic algorithm for the atemporal case, we referred the solution to the decision-making agent. In this case we want to be more precise.

**Definition 8** *Let $\langle \mathbb{A}, \prec, \chi \rangle$ be a basic model for recovery actions. We define the function* **merge** $: 2^{\mathbb{A}} \mapsto 2^{\mathbb{A}}$ *as follows:*

**(4)** $$\mathbf{merge}(S) = \{a_i \in S \mid \text{there is no } a_j \in S \text{ such that } a_i \prec a_j\}$$

We moreover define:

$$(5) \qquad \mathbf{merge\text{-}set}(\mathbb{A}) = \{\mathbf{merge}(S) \mid S \in 2^{\mathbb{A}}\} \subseteq 2^{\mathbb{A}}$$

$\mathbf{merge\text{-}set}(\mathbb{A})$ is the set of *compound recovery actions* which includes basic recovery actions in the form of singletons. $\diamond$

The intuition behind **merge** is that when we cannot discriminate multiple fault situations we simply merge the corresponding recovery actions. This means that we collect all recovery actions, and then remove the unnecessary ones (equation (4)). An action in a set becomes unnecessary when the set contains a stronger action. Thus given a te-set $\mathbb{E}$, we define:

$$(6) \qquad \mathbf{Act}(\mathbb{E}) = \mathbf{merge}(\{\mathbf{Act}(\mathbf{sit}) \mid \mathbf{sit} \in \mathbb{E}\}).$$

If we take into account compound actions, we can extend the notion of model for recovery actions as follows:

**Definition 9** Let $\mathbf{A} = \langle \mathbb{A}, \prec, \chi \rangle$ be a basic model for recovery actions. An *extended model* for $\mathbf{A}$ is a triple $\langle \mathbb{A}, \prec_{\mathbf{ext}}, \chi_{\mathbf{ext}} \rangle$ where:

**(1)** $\prec_{\mathbf{ext}} \subseteq \mathbf{merge\text{-}set}(\mathbb{A}) \times \mathbf{merge\text{-}set}(\mathbb{A})$ and given $A, A' \in \mathbf{merge\text{-}set}(\mathbb{A})$, $A \prec_{\mathbf{ext}} A'$ if for every $a \in A$ either $a \in A'$ or there exists $a' \in A'$ such that $a \prec a'$. Notice that $\{a_i\} \prec_{\mathbf{ext}} \{a_j\}$ if and only if $a_i \prec a_j$.

**(2)** $\chi_{\mathbf{ext}} : \mathbf{merge\text{-}set}(\mathbb{A}) \mapsto \mathbb{R}^+$ is a cost function over compound actions such that for every $A \in \mathbf{merge\text{-}set}(\mathbb{A})$, $\max_{a \in A} \chi(a) \leq \chi_{\mathbf{ext}}(A) \leq \sum_{a \in A}$. Moreover, if $A \prec_{\mathbf{ext}} A'$ then it must hold that $\chi_{\mathbf{ext}}(A) < \chi_{\mathbf{ext}}(A')$.

While $\prec_{\mathbf{ext}}$ is uniquely determined by $\prec$, the same does not hold for $\chi_{\mathbf{ext}}$: for this reason there is more than one extended model for any basic model. The requirement that $\max_{a \in A} \chi(a) \leq \chi_{\mathbf{ext}}(A)$ is motivated as follows: if there existed $a \in A$ such that $\chi_{\mathbf{ext}}(A) \leq \chi(a)$ then it would make sense never to perform $a$, substituting it for $A$. In fact, $\{a\} \prec_{\mathbf{ext}} A$ and $A$ would have the same or lower cost. We also ask $\chi_{\mathbf{ext}}$ - as we did for basic models - to be monotonic with respect to $\prec_{\mathbf{ext}}$.

In the following we shall consider only extended models for recovery actions, thus we shall drop the **ext** prefix from both $\prec$ and $\chi$.

***Example 4*** *Figure 7 shows a possible extension for the basic model in Figure 6. In this case the cost of the compound action $\{b, c\}$ is given by the sum of the individual costs of $b$ and $c$.* $\diamond$

## 5. The Problem of Building Temporal Decision Trees

In this section we outline the peculiarities of building temporal decision trees, showing the differences with respect to the "traditional" case.

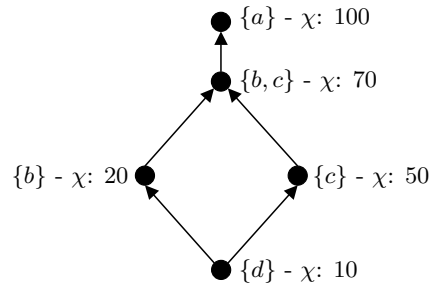

Figure 7: An extended action model.

## 5.1 The Challenge of Temporal Decision Trees

What makes the generation of temporal decision trees more difficult from standard ones is the requirement that time labels do not decrease when moving from the root to the leaves: this corresponds to assuming that sensor values cannot be stored; when the decision-making agent decides to wait it gives up using all values that sensors show while it is waiting.

If we release this restriction we can actually generate temporal decision trees with a minor variation of **ID3**, essentially by considering each pair formed by a sensor and a time label as an individual test. In other words ID3CHOOSETEST selects a sensor $s$ and a time label $t$ such that reading $s$ at time $t$ allows for the maximum discrimination among examples.

However in systems as the ones we are considering, that is low-memory real-time systems, the possibility of performing the diagnostic task without discarding dynamics but also without having to store sensor values across time is a serious issue to take into account. For this reason the definition of temporal decision tree includes the requirement that time labels be not decreasing on root-leaf paths.

Figure 8 shows a generic algorithm for building temporal decision trees that can help us outline the difficulties of the task. Line 8 shows a minor modification aimed at taking into account deadlines: an observation can be used on a given set of examples only if its time label does not exceed its global deadline. Violating this condition would result in a tree that selects a recovery action only after the deadline for the corresponding fault situation has expired.

The major change with respect to the standard algorithm is however shown in line 15: once we select an observation pair ⟨`sensor`,`tlabel`⟩ we must remove from the set of observations all those pairs whose time label is lower than `tlabel`[5].

As a consequence of these operations - ruling out invalid observations and discarding those that are in the past - the set of observations available when building a child node can be different from the one used for its parent in more than one way:

---

```
1    function BuildTemporalTree (te-set Examples, set Obs, action model ActModel)
2    returns a temporal decision tree T = ⟨root, Nodes, Edges, Labels, TLabels⟩
3    begin
4        if for all sit ∈ Examples Act(sit) is the same then
5            return BuildLeaf(Examples, ActModel);
6        deadline ← Dl(Examples);
7        UsefulObs ← {o ∈ Obs | ∃s1, s2 ∈ Examples s.t. Val(s1, o) ≠ Val(s2, o)};
8        ValidObs ← {⟨sensor, tlabel⟩ ∈ UsefulObs | tlabel ≤ deadline};
9        if ValidObs is empty then
10           return BuildLeaf(Examples, ActModel);
11       ⟨sensor, tlabel⟩ ← ChooseObs (ValidObs, Examples, ActModel);
12       root ← new node;
13       Nodes ← {root}; Edges ← ∅; Labels(root) ← sensor; TLabels(root) ← tlabel;
14       T ← ⟨root, Nodes, Edges, Labels, TLabels⟩;
15       Obs_Update ← {⟨sens, tst⟩ ∈ UsefulObs | tst ≥ tlabel};
16       for each possible measure value of sensor do begin
17           SubExamples ← {sit ∈ Examples | Val(sit, ⟨sensor, tlabel⟩) = value};
18           if SubExamples is not empty then begin
19               SubTree ← BuildTemporalTree (SubExamples, Obs_Update, ActModel);
20               Append(T, root, SubTree);
21               Labels((root, Root(SubTree))) ← value;
22           end;
23       end;
24       return T;
25   end.

26   function BuildLeaf (te-set Examples, action model ActModel)
27   returns a loose temporal decision tree T = ⟨leaf, {leaf}, ∅, Labels, ∅⟩
28   begin
29       all_act ← {Act(sit) | sit ∈ Examples};
30       comp_act ← merge(all_act);
31       leaf ← new node; Labels(leaf) ← comp_act; T ← ⟨leaf, {leaf}, ∅, Labels, ∅⟩;
32       return T;
33   end.
```

Figure 8: Generic algorithm for building a temporal decision tree

- Some observations can be invalid for the parent and valid for the child. The recursive call for the child works on a smaller set of examples; therefore the global deadline may be further ahead in time, allowing more observations to be used.

- Some observations can become unavailable for the child because they have a time label lower than that used for the parent.

Of course the problematic issue is the latter: some observations are lost, and among them there may be some information which is *necessary* for properly selecting a recovery action.

Let us consider as an example the te-set in Figure 9, with four fault situations, two time labels (0 and 1) and only one sensor ($s$). Each fault situation is characterized by a different recovery action, and the te-set obviously allows to discriminate all of them. However the entropy criterion would first select the observation $\langle s, 1 \rangle$, which is more discriminating.

|                    | $\langle s,0 \rangle$ | $\langle s,1 \rangle$ | **Act** |
|--------------------|------|------|-----|
| **sit$_1$**        | x    | x    | a   |
| **sit$_2$**        | x    | y    | b   |
| **sit$_3$**        | y    | y    | c   |
| **sit$_4$**        | y    | z    | d   |

Figure 9: A te-set causing some problems to standard **ID3** algorithm if used for temporal decision trees.

The observation $\langle s,0 \rangle$ would then become unavailable, and the resulting tree could never discriminate **sit$_2$** and **sit$_3$**.

This shows that there is a relevant difference between building standard decision trees and building temporal decision trees. Let us look again at the generic algorithm for standard decision trees presented in Figure 3: the particular strategy implemented in CHOOSETEST does not affect the capability of the tree of selecting the proper recovery action, but only the size of the tree. Essentially the tree contains the same information as the set of examples - at least for what concerns the association between observations and recovery actions. We can say that *the tree has always the same discriminating power as the set of examples*, meaning that the only case when the tree is not capable of deciding between two recovery actions is when the set of examples contains two fault situations with identical observations and different actions.

If we consider the algorithm in Figure 8 we see that the order in which observations are selected - that is, the particular implementation of CHOOSEOBS - can affect the discriminating power of the tree, and not only its size. Since from one recursive call to the following some observations are discarded, we can obtain a tree with *less discriminating information than the original set of examples*. Our primary task is then to avoid such a situation, that is to build a tree which is small, but which does not sacrifice relevant information. As a consequence, we cannot exploit the strategy of simply selecting an observation with minimum entropy.

In the next sections we shall formalize the new requirements for the output tree, and propose an implementation of CHOOSEOBS which meets them.

## 5.2 Each Tree Has a Cost

In the previous section we informally introduced the notion of discriminating power. In this section we shall introduce a more general notion of *expected cost* of a temporal decision tree. Intuitively, the expected cost associated with a temporal decision tree is the expected cost of a recovery action selected by the tree, with respect to the probability distribution of the fault situations.

Expected cost is a stronger notion than discriminating power: on the one hand if a tree discriminates better than another, than it has also a lower expected cost (we shall soon prove this statement). On the other hand expected cost adds something to the notion of discriminating power, since any two trees are comparable from the point of view of cost, while they may not be from the point of view of the discrimination they carry out.

Before defining expected cost, we need to introduce some preliminary definitions.

We shall make use of a function, named **examples**, that given an initial set of examples $\mathbb{E}$ and a tree **T** associates to each node of the tree a subset of $\mathbb{E}$. To understand the meaning of such a function before formally defining it, let us imagine to run the tree on $\mathbb{E}$ and at a certain point of the decision process to reach a node $n$: **examples** tells us which is the subset of fault situations which we have not yet discarded.

**Definition 10** Let $\mathbb{E}$ be a te-set with sensors $s_1, \ldots, s_m$, time labels $t_1, \ldots, t_{\mathsf{last}}$ and actions model $\langle \mathbb{A}, \prec, \chi \rangle$. Moreover let $\mathbf{T} = \langle r, N, E, \mathcal{L}, \mathcal{T} \rangle$ be a temporal decision tree such that for every internal node $n \in N$ we have $\mathcal{L}(n) \in \{s_1, \ldots, s_m\}$ and $\mathcal{T}(n) \in \{t_1, \ldots, t_{\mathsf{last}}\}$. We define a function $\mathbf{examples}(\cdot; \mathbb{E}) : N \mapsto 2^{\mathbb{E}}$ as follows:

(7)  $\mathbf{examples}(r; \mathbb{E}) = \mathbb{E}$  where $r$ is the root of **T**;

  $\mathbf{examples}(n; \mathbb{E}) = \{\mathbf{sit} \in \mathbf{examples}(p; \mathbb{E}) \mid \mathbf{Val}(\mathbf{sit}, \langle \mathcal{L}(p), \mathcal{T}(p) \rangle) = \mathcal{L}((p, n))\}$

  where $n \in N$, $n \neq r$ and $(p, n) \in E$.

**examples** is well defined since for any $n \in N$ different from the root there exists exactly one $p \in N$ such that $(p, n) \in E$.  $\diamond$

Notice that, if $\mathbb{E}$ is the set of examples used for building **T**, $\mathbf{examples}(n; \mathbb{E})$ corresponds to the subset of examples used while creating node $n$.

**Example 5** *Let us consider Figure 10: it shows the same tree as Figure 5, but for every node $n$ we can also see the set of fault situations $\mathbf{examples}(n; \mathbb{E})$, where $\mathbb{E}$ is the te-set of Figure 1.*  $\diamond$

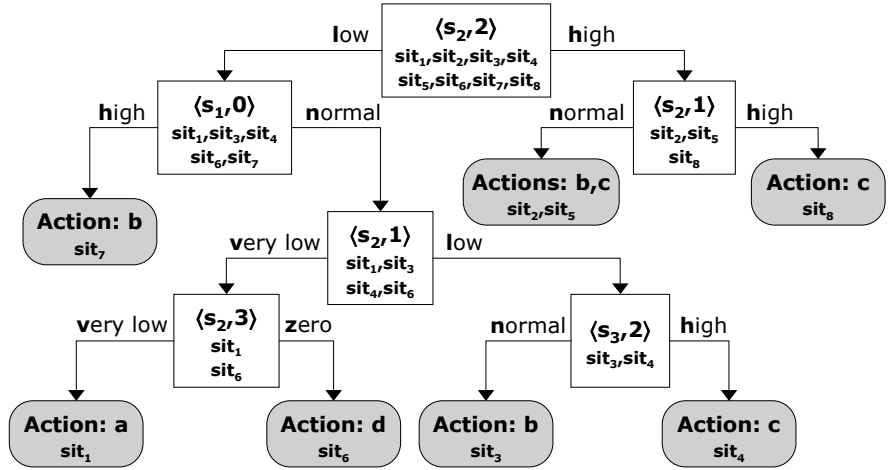

Figure 10: A temporal decision tree showing the value of $\mathbf{example}(n, \mathbb{E})$.

In the following when using function **examples** we shall omit the second argument, denoting the initial te-set, when there is no ambiguity about which te-set is considered.

Not every tree can be used on a given set of examples: actually we need some compatibility between the two, which is characterized by the following definition.

**Definition 11** Let $\mathbb{E}$ be as in previous definition. We say that a temporal decision tree $T = \langle r, N, E, \mathcal{L}, \mathcal{T} \rangle$ is *compatible with* $\mathbb{E}$ if:

- For every internal node $n \in N$, $\mathcal{L}(n) \in \{s_1, \ldots, s_m\}$, $\mathcal{T}(n) \in \{t_1, \ldots, t_{\textbf{last}}\}$ and $\mathcal{T}(n) \leq$ **DI**(**examples**$(n; \mathbb{E})$).

- For every leaf $l \in N$, $\mathcal{L}(l) \in$ **merge**$(\mathbb{A})$ and $\mathcal{L}(l) =$ **merge**(**Act**(**examples**$(l; \mathbb{E})$)).

It is straightforward to see that a tree is compatible with the set of examples used in building it.

We have the following property[6]:

**Proposition 12** *Let* $\textbf{T} = \langle r, N, E, \mathcal{L}, \mathcal{T} \rangle$ *be a temporal decision tree compatible with a te-set* $\mathbb{E}$. *Let* $l_1, \ldots, l_f \in N$ *denote the leaves of* $T$. *Then* **examples**$(l_1), \ldots,$ **examples**$(l_f)$ *is a partition of* $\mathbb{E}$.

For each $\textbf{sit} \in \mathbb{E}$ we then denote by $\textbf{leaf}_\textbf{T}(\textbf{sit})$ the unique leaf $l$ of $\textbf{T}$ such that $\textbf{sit} \in$ **examples**$(l)$. We are now ready to formalize the notion of discriminating power.

**Definition 13** Let $\textbf{T} = \langle r_\textbf{T}, N_\textbf{T}, E_\textbf{T}, \mathcal{L}_\textbf{T}, \mathcal{T}_\textbf{T} \rangle$, $\textbf{U} = \langle r_\textbf{U}, N_\textbf{U}, E_\textbf{U}, \mathcal{L}_\textbf{U}, \mathcal{T}_\textbf{U} \rangle$ denote two temporal decision trees compatible with the same te-set $\mathbb{E}$. Let moreover $\langle \mathbb{A}, \prec, \chi \rangle$ be the recovery action model used in building $\textbf{T}$ and $\textbf{U}$. We say that $\textbf{T}$ is *more discriminating than* $\textbf{U}$ *with respect to* $\mathbb{E}$ if:

**(1)** for every $\textbf{sit} \in \mathbb{E}$ either $\mathcal{L}_\textbf{T}(\textbf{leaf}_\textbf{T}(\textbf{sit})) \prec \mathcal{L}_\textbf{U}(\textbf{leaf}_\textbf{U}(\textbf{sit}))$ or $\mathcal{L}_\textbf{T}(\textbf{leaf}_\textbf{T}(\textbf{sit})) = \mathcal{L}_\textbf{U}(\textbf{leaf}_\textbf{U}(\textbf{sit}))$;

**(2)** there exists $\textbf{sit} \in \mathbb{E}$ such that $\mathcal{L}_\textbf{T}(\textbf{leaf}_\textbf{T}(\textbf{sit})) \prec \mathcal{L}_\textbf{U}(\textbf{leaf}_\textbf{U}(\textbf{sit}))$.　　　　　◇

Notice that the second condition makes sure that the two trees are not equal (in which case they would be *equally discriminating*), something that the first condition alone cannot guarantee.

**Example 6** *Let us consider the tree in Figure 10 above and the tree in Figure 11 below. The former is more discriminating than the latter. In fact, the two trees associate the same actions to* $\textbf{sit}_1, \textbf{sit}_2, \textbf{sit}_3, \textbf{sit}_4, \textbf{sit}_5$ *and* $\textbf{sit}_6$. *However the former associates action b with* $\textbf{sit}_7$ *and action c with* $\textbf{sit}_8$, *while the latter associates to both* $\textbf{sit}_7$ *and* $\textbf{sit}_8$ *the compound action* $\{b, c\}$.　　　　　◇

Unfortunately we cannot easily use discriminating power - as defined above - as a preference criterion for decision trees. The reason is that it does not define a *total* order on decision trees, but only a partial one. The following situations may arise:

---

6. For the sake of readability, all proofs are collected in a separate appendix at the end of the paper.

- For some **sit**, $\mathcal{L}_{\mathbf{T}}(\mathbf{leaf_T}(\mathbf{sit})) \prec \mathcal{L}_{\mathbf{U}}(\mathbf{leaf_U}(\mathbf{sit}))$; for some other **sit**, $\mathcal{L}_{\mathbf{U}}(\mathbf{leaf_U}(\mathbf{sit})) \prec \mathcal{L}_{\mathbf{T}}(\mathbf{leaf_T}(\mathbf{sit}))$.

- For a given **sit**, $\mathcal{L}_{\mathbf{T}}(\mathbf{leaf_T}(\mathbf{sit})) \not\prec \mathcal{L}_{\mathbf{U}}(\mathbf{leaf_U}(\mathbf{sit}))$, nor $\mathcal{L}_{\mathbf{U}}(\mathbf{leaf_U}(\mathbf{sit})) \not\prec \mathcal{L}_{\mathbf{T}}(\mathbf{leaf_T}(\mathbf{sit}))$.

From the point of view of discriminating power alone, it is reasonable for **T** and **U** not to be comparable in the above cases. Nonetheless, there may be a reason for preferring one over the other, and this reason is *cost*. For example if we consider the second situation, even if $\mathcal{L}_{\mathbf{T}}(\mathbf{sit})$ and $\mathcal{L}_{\mathbf{U}}(\mathbf{sit})$ are not directly comparable from the point of view of their strength, one of the two may be cheaper than the other and thus preferable.

We therefore introduce the notion of expected cost of a tree.

**Definition 14** Let $\mathbf{T} = \langle r, N, E, \mathcal{L}, \mathcal{T} \rangle$ be a temporal decision tree compatible with a te-set $\mathbb{E}$ and an action model $\mathbf{A} = \langle \mathbb{A}, \prec, \chi \rangle$. We inductively define an *expected cost* function $\mathcal{X}_{\mathbb{E}, \mathbf{A}} : N \mapsto \mathbb{R}^+$ on tree nodes as follows:

$$(8) \qquad \mathcal{X}_{\mathbb{E}, \mathbf{A}}(n) = \begin{cases} \chi(\mathcal{L}(l)) & \text{if } l \in N \text{ is a leaf;} \\ \displaystyle\sum_{c:(n,c) \in E} P(\mathcal{L}(n) \to \mathcal{L}((n,c))) \cdot \mathcal{X}_{\mathbb{E}, \mathbf{A}}(c) & \text{if } n \in N \text{ is an inner node.} \end{cases}$$

where $P(\mathcal{L}(n) \to \mathcal{L}((n,c)))$ is the probability of sensor $\mathcal{L}(n)$ showing a value $v = \mathcal{L}((n,c))$ and is given by:

$$P(\mathcal{L}(n) \to \mathcal{L}((n,c))) = \frac{P(\mathbf{examples}(c); \mathbb{E})}{P(\mathbf{examples}(n); \mathbb{E})} = P(\mathbf{examples}(c); \mathbf{examples}(n)).$$

The *expected cost of* **T** *with respect to* $\mathbb{E}$ *and* **A**, denoted by $\mathcal{X}_{\mathbb{E}, \mathbf{A}}(\mathbf{T})$, is then defined as:

$$(9) \qquad \mathcal{X}_{\mathbb{E}, \mathbf{A}}(\mathbf{T}) = \mathcal{X}_{\mathbb{E}, \mathbf{A}}(r) \quad \text{where } r \text{ is the root of } \mathbf{T}$$

$\diamond$

The above definition states that:

- The expected cost of a tree leaf $l$ is simply the cost of its recovery action $\mathcal{L}(l)$;

- The expected cost of an inner node $n$ is given by the weighted sum of its children's expected costs; weight for child $c$ is given by the probability $P(\mathcal{L}(n) \to \mathcal{L}((n,c)))$.

- The expected cost of a temporal decision tree **T** is the expected cost of its root.

The following proposition states that the weighted sum for computing the expected cost of the root can be performed directly on tree leaves.

**Proposition 15** Let $\mathbf{T} = \langle r, N, E, \mathcal{L}, \mathcal{T} \rangle$ denote a temporal decision tree, and let $l_1, \ldots, l_u$ be its leaves. Then

$$(10) \qquad \mathcal{X}_{\mathbb{E}, \mathbf{A}}(\mathbf{T}) = \sum_{i=1}^{u} \chi(\mathcal{L}(l_i)) \cdot P(\mathbf{examples}(l); \mathbb{E})$$

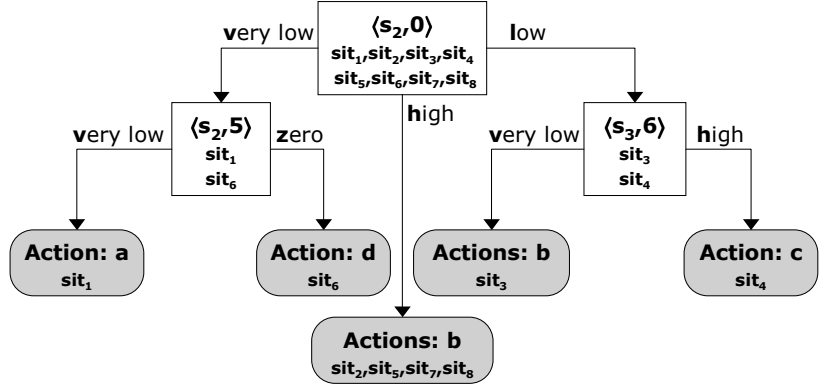

Figure 11: A temporal decision tree less discriminating than the one in Figure 10.

The next proposition shows that expected cost is monotonic with respect to the "better discrimination" relation, and therefore it is a good preference criterion for temporal decision trees, since a tree with the lowest possible expected cost is the most discriminating one, and moreover it is the cheapest among equally discriminating trees.

**Proposition 16** *Let* $\mathbf{T} = \langle r_\mathbf{T}, N_\mathbf{T}, E_\mathbf{T}, \mathcal{L}_\mathbf{T}, \mathcal{T}_\mathbf{T} \rangle$, $\mathbf{U} = \langle r_\mathbf{U}, N_\mathbf{U}, E_\mathbf{U}, \mathcal{L}_\mathbf{U}, \mathcal{T}_\mathbf{U} \rangle$ *be two temporal decision trees compatible with the same te-set* $\mathbb{E}$ *and the same actions model* $\mathbf{A}$. *If* $\mathbf{T}$ *is more discriminating than* $\mathbf{U}$ *then* $\mathcal{X}_{\mathbb{E},\mathbf{A}}(\mathbf{T}) < \mathcal{X}_{\mathbb{E},\mathbf{A}}(\mathbf{U})$.

**Example 7** *Let us compute the expected cost of tree* $\mathbf{T}_1$ *in Figure 10 and of tree* $\mathbf{T}_2$ *in Figure 11, with respect to the te-set* $\mathbb{E}$ *in Figure 1 and the action model* $\mathbf{A}$ *in Figure 7. We shall assume that all fault situations are equiprobable, that is each of them has probability 1/8. By exploiting proposition 15 we obtain:*

$$
\begin{aligned}
\mathcal{X}_{\mathbb{E},\mathbf{A}}(\mathbf{T}_1) &= P(\mathbf{sit}_1)\chi(a) + P(\mathbf{sit}_7)\chi(b) + P(\mathbf{sit}_6)\chi(d) + P(\mathbf{sit}_3)\chi(b) + P(\mathbf{sit}_4)\chi(c) + \\
&\quad + P(\{\mathbf{sit}_2, \mathbf{sit}_5\})\chi(\{b, c\}) + P(\mathbf{sit}_8)\chi(c) = \\
&= \frac{1}{8} \cdot 100 + \frac{1}{8} \cdot 20 + \frac{1}{8} \cdot 10 + \frac{1}{8} \cdot 20 + \frac{1}{8} \cdot 50 + \frac{1}{4} \cdot 70 + \frac{1}{8} \cdot 50 = \\
&= 12.5 + 2.5 + 1.25 + 2.5 + 6.25 + 17.5 + 6.25 = 48.75 \\
\mathcal{X}_{\mathbb{E},\mathbf{A}}(\mathbf{T}_2) &= P(\mathbf{sit}_1)\chi(a) + P(\mathbf{sit}_6)\chi(d) + P(\mathbf{sit}_3)\chi(b) + P(\mathbf{sit}_4)\chi(c) + \\
&\quad + P(\{\mathbf{sit}_2, \mathbf{sit}_5, \mathbf{sit}_7, \mathbf{sit}_8\})\chi(\{b, c\}) = \\
&= \frac{1}{8} \cdot 100 + \frac{1}{8} \cdot 10 + \frac{1}{8} \cdot 20 + \frac{1}{8} \cdot 50 + \frac{1}{2} \cdot 70 = \\
&= 12.5 + 1.25 + 2.5 + 6.25 + 35 = 57.5
\end{aligned}
$$

*We can see that the less discriminating tree, that is* $\mathbf{T}_2$, *has a higher expected cost.* ◇

### 5.3 Restating the Problem

In the previous section we introduced expected cost as a preference criterion for decision trees. Given this notion, we can restate the problem of building temporal decision tree as that of building a tree with minimum possible expected cost. This section formally shows that the notion of "minimum possible expected cost" is well defined, and more precisely it corresponds to the cost of a tree that exploits all observations in the te-set. The goal can then be expressed as *finding a reasonably small tree among those whose expected cost is minimum*.

In this section, as well as formalizing the above mentioned notions, we introduce some formal machinery that will be useful in proving the correctness of our algorithm.

**Definition 17** Let $\mathbb{E}$ denote a te-set. Moreover, let $t_1, \ldots, t_{\text{last}}$ denote the time labels of $\mathbb{E}$. We say that $\mathbf{sit}_i, \mathbf{sit}_j \in \mathbb{E}$ are *pairwise indistinguishable*, and we write $\mathbf{sit}_i \sim \mathbf{sit}_j$, if for all $t_i < \min\{\mathbf{Dl}(\mathbf{sit}_i), \mathbf{Dl}(\mathbf{sit}_j)\}$ and for all sensors $s$ we have that $\mathbf{Val}(\mathbf{sit}_i, \langle s, t_i \rangle) = \mathbf{Val}(\mathbf{sit}_j, \langle s, t_i \rangle)$. ◇

As a relation, $\sim$ is obviously reflexive and symmetric, but it is not transitive. If we consider a $\mathbf{sit}_k$ with a particularly strict deadline, it might well be that $\mathbf{sit}_i \sim \mathbf{sit}_k$, $\mathbf{sit}_k \sim \mathbf{sit}_j$, but $\mathbf{sit}_i \nsim \mathbf{sit}_j$. We now introduce a new relation $\approx$ which is the transitive closure of $\sim$.

**Definition 18** Let $\mathbb{E}$ denote a te-set. We say that $\mathbf{sit}_i, \mathbf{sit}_j \in \mathbb{E}$ are *indistinguishable*, and we write $\mathbf{sit}_i \approx \mathbf{sit}_j$, if there exists a finite sequence $\mathbf{sit}_{k_1}, \ldots, \mathbf{sit}_{k_u} \in \mathbb{E}$ such that

- $\mathbf{sit}_{k_1} = \mathbf{sit}_i$;
- $\mathbf{sit}_{k_u} = \mathbf{sit}_j$;
- for every $g = 1, \ldots, u - 1$, $\mathbf{sit}_{k_g} \sim \mathbf{sit}_{k_{g+1}}$. ◇

$\approx$ is an equivalence relation over $\mathbb{E}$, and we denote by $\mathbb{E}/\approx$ the corresponding quotient set. We have the following definition:

**Definition 19** Let $\mathbb{E}$ be a te-set, with actions model $\mathbf{A}$. The *expected cost of $\mathbb{E}$*, denoted by $\overline{\mathcal{X}}_{\mathbb{E}, \mathbf{A}}$, is defined as:

$$(11) \qquad \overline{\mathcal{X}}_{\mathbb{E}, \mathbf{A}} = \sum_{\eta \in \mathbb{E}/\approx} \chi(\mathbf{merge}(\{\mathbf{Act}(\mathbf{sit}) \mid \mathbf{sit} \in \eta\})) \cdot P(\eta; \mathbb{E})$$

◇

**Example 8** *Let us consider the te-set $\mathbb{E}$ in Figure 1 and the action model $\mathbf{A}$ in Figure 7. The only two indistinguishable fault situations in $\mathbb{E}$ are $\mathbf{sit}_2$ and $\mathbf{sit}_5$. Thus we have:*

$$\begin{aligned}
\overline{\mathcal{X}}_{\mathbb{E}, \mathbf{A}} &= P(\mathbf{sit}_1)\chi(a) + P(\{\mathbf{sit}_2, \mathbf{sit}_5\})\chi(\{b, c\}) + P(\mathbf{sit}_3)\chi(b) + P(\mathbf{sit}_4)\chi(c) + \\
&\quad + P(\mathbf{sit}_6)\chi(d) + P(\mathbf{sit}_7)\chi(b) + P(\mathbf{sit}_8)\chi(c) = \\
&= \frac{1}{8} \cdot 100 + \frac{1}{4} \cdot 70 + \frac{1}{8} \cdot 20 + \frac{1}{8} \cdot 50 + \frac{1}{8} \cdot 10 + \frac{1}{8} \cdot 20 + \frac{1}{8} \cdot 50 = \\
&= 12.5 + 17.5 + 2.5 + 6.25 + 1.25 + 2.5 + 6.25 = 48.75
\end{aligned}$$

*Notice that the tree in Figure 10 has the same cost as the te-set, thus its cost is the minimum possible, as we show below. Of course we may still be able to build another smaller tree with the same cost.* ◇

Now we need to show that $\overline{\mathcal{X}}_{\mathbb{E},\mathbf{A}}$ is actually the minimum possible expected cost for a temporal decision tree compatible with $\mathbb{E}$.

**Theorem 20** *Let $\mathbb{E}$ be a te-set with actions model $\mathbf{A}$. We have that:*

(i) *There exists a decision tree $\overline{\mathbf{T}}$ compatible with $\mathbb{E}$ such that $\mathcal{X}_{\mathbb{E},\mathbf{A}}(\overline{\mathbf{T}}) = \overline{\mathcal{X}}_{\mathbb{E},\mathbf{A}}$.*

(ii) *For every temporal decision tree $\mathbf{T}$ compatible with $\mathbb{E}$, $\overline{\mathcal{X}}_{\mathbb{E},\mathbf{A}} \leq \mathcal{X}_{\mathbb{E},\mathbf{A}}(\mathbf{T})$.* ◇

Now we can state more precisely the problem of building a temporal decision tree:

> *Given a te-set $\mathbb{E}$ with actions model $\mathbf{A}$, we want to build a temporal decision tree $\mathbf{T}$ over $\mathbb{E}$, such that $\mathcal{X}_{\mathbb{E},\mathbf{A}}(\mathbf{T}) = \overline{\mathcal{X}}_{\mathbb{E},\mathbf{A}}$. Moreover, we want to keep the tree reasonably small by exploiting entropy.*

## 6. The Algorithm

In this section we describe in detail our proposal for building temporal decision trees from a given te-set and action model. We also discuss the complexity of the algorithm we introduce, and give an example of how the algorithm works.

### 6.1 Preconditions

Our goal is now to define an implementation of function CHOOSEOBS such that, once plugged into function BUILDTEMPORALTREE, yields a solution to the problem of building temporal decision trees as stated in section 5.3. First however we shall analyze some properties of BUILDTEMPORALTREE as defined in Figure 8: this will lead us smoothly to the solution and will help us prove formally its correctness. In order to accomplish this task we need to introduce some notation that allows us to speak about algorithm properties.

Let $\mathbb{E}$ be a te-set with fault situations $\{\mathbf{sit}_1, \ldots, \mathbf{sit}_n\}$, sensors $\{s_1, \ldots, s_m\}$, time labels $\{t_1, \ldots, t_{\mathbf{last}}\}$ and action model $\mathbf{A}$. We aim at computing our tree $\mathbf{T}$ by executing:

(12)    $\mathbf{T} \leftarrow \text{BUILDTEMPORALTREE}(\{\mathbf{sit}_1, \ldots, \mathbf{sit}_n\}, \{s_1, \ldots, s_m\} \times \{t_1, \ldots, t_{\mathbf{last}}\}, \mathbf{A})$

Each execution of BUILDTEMPORALTREE comprises several recursive calls to the same function; given two recursive calls $c, c'$ we shall write $c \sqsubset c'$ when $c$ occurs immediately inside $c'$. Moreover we shall denote by $c_0$ the *initial* call. Finally, we shall call *terminal* a recursive call which does not have any further inner call.

For a given call $c$, we shall denote by $[\![\texttt{Example}]\!]_c$, $[\![\texttt{Obs}]\!]_c$, $[\![\texttt{ActModel}]\!]_c$ the actual values of the formal parameters in $c$. With a slight abuse of notation, we shall also write $[\![\texttt{var}]\!]_c$ to denote the value of those variables $\texttt{var}$ in $c$ that, once set, never change their value ($\texttt{deadline}$, $\texttt{UsefulObs}$, $\texttt{ValidObs}$, $\texttt{sensor}$, $\texttt{tlabel}$, $\texttt{Obs\_Update}$). Finally, we shall denote by $[\![\mathbf{T}]\!]_c$ the tree returned by call $c$.

Each recursive call $c$ works on a different te-set, which is defined by $[\![\texttt{Examples}]\!]_c$ and $[\![\texttt{Obs}]\!]_c$. The actions model however is always the same, since for $c \sqsubset c'$ we have $[\![\texttt{ActModel}]\!]_c = [\![\texttt{ActModel}]\!]_{c'}$. We shall denote by $\mathbb{E}_c$ the te-set used in call $c$, which is determined by its input parameters.

The following proposition is critical for proving the correctness of our approach. It states that we can obtain a tree with minimum expected cost if and only if we guarantee that there is no increase in the expected cost of the te-set when passing from the set of observations Obs to the set Obs_Update:

**Proposition 21** *Let us consider an execution of* BuildTemporalTree *starting with a main call $c_0$. The initial te-set, which we want to build a tree over, is $\mathbb{E} = \mathbb{E}_{c_0}$ with $\mathbf{A} = [\![\texttt{ActModel}]\!]_{c_0}$. For any recursive call $c$, let us denote by $\mathbb{E}_c^*$ the te-set determined by $[\![\texttt{Examples}]\!]_c$ and $[\![\texttt{Obs\_Update}]\!]_c$. Then:*

**(1)** $\mathcal{X}_{\mathbb{E},\mathbf{A}}([\![\texttt{T}]\!]_{c_0}) \geq \overline{\mathcal{X}}_{\mathbb{E},\mathbf{A}}$

**(2)** $\mathcal{X}_{\mathbb{E},\mathbf{A}}([\![\texttt{T}]\!]_{c_0}) = \overline{\mathcal{X}}_{\mathbb{E},\mathbf{A}}$ *if and only if for every non terminal[7] recursive call $c$ generated by $c_0$ it holds that $\overline{\mathcal{X}}_{\mathbb{E}_c,\mathbf{A}} = \overline{\mathcal{X}}_{\mathbb{E}_c^*,\mathbf{A}}$*

### 6.2 Implementing ChooseObs

Proposition 21 suggests that we need to provide an implementation for ChooseObs such that at each recursive call $c$, $\overline{\mathcal{X}}_{\mathbb{E}_c,\mathbf{A}} = \overline{\mathcal{X}}_{\mathbb{E}_c^*,\mathbf{A}}$. Let us examine in more detail the relations between $[\![\texttt{Obs}]\!]_c$ and $[\![\texttt{Obs\_Update}]\!]_c$.

As a first step, $[\![\texttt{UsefulObs}]\!]_c$ is obtained by removing from $[\![\texttt{Obs}]\!]_c$ those observations that do not help in discriminating fault situations. This has no effect on the expected cost of the te-set, since it does not affect the relation of indistinguishability.

Then $[\![\texttt{Obs\_Update}]\!]_c$ is obtained from $[\![\texttt{UsefulObs}]\!]_c$ by removing those observations whose time label precedes the chosen one. The expected cost of the resulting te-set thus depends on the time label selected by function ChooseObs. We have the following properties:

**Proposition 22** *Let $c, d$ denote two independent calls to* BuildTemporalTree *with the same input arguments but with different implementations of* ChooseObs. *If $[\![\texttt{tlabel}]\!]_c \leq [\![\texttt{tlabel}]\!]_d$ then $\overline{\mathcal{X}}_{\mathbb{E}_c^*,\mathbf{A}} \leq \overline{\mathcal{X}}_{\mathbb{E}_d^*,\mathbf{A}}$.*

**Proposition 23** *Let $c$ be a call to* BuildTemporalTree. *If $[\![\texttt{tlabel}]\!]_c = t_{\mathbf{min}_c} = \min\{t \mid \langle t,s \rangle \in [\![\texttt{ValidObs}]\!]_c\}$ then $\overline{\mathcal{X}}_{\mathbb{E}_c^*,\mathbf{A}} \leq \overline{\mathcal{X}}_{\mathbb{E}_c,\mathbf{A}}$.*

Now we define the notion of a *safe* time label:

**Definition 24** Let $c$ denote a call to BuildTemporalTree. A time label $t$ is said to be *safe with respect to $c$* if $[\![\texttt{tlabel}]\!]_c = t$ implies $\overline{\mathcal{X}}_{\mathbb{E}_c^*,\mathbf{A}} = \overline{\mathcal{X}}_{\mathbb{E}_c,\mathbf{A}}$.

An immediate consequence of propositions 22 and 23 is the following:

**Proposition 25** *For any call $c$ to* BuildTemporalTree *there exist a time label $t_{\mathbf{max}_c}$ such that the safe time labels are all and only those $t$ with $t_{\mathbf{min}_c} \leq t \leq t_{\mathbf{max}_c}$, where $t_{\mathbf{min}_c}$ is as in proposition 23.*

---

7. We exclude terminal calls because they do not even compute Obs_Update.

Figure 12 describes ID3ChooseSafeObs, the implementation of ChooseObs we propose. It exploits the properties we have proved in this and the previous section in order to achieve the desired task in an efficient way. Let us examine it in more detail.

ID3ChooseSafeObs (Figure 12) computes the set of safe observations (line 4) and then chooses among them one with minimum entropy (line 5). For what we have proved up to now, such an implementation yields a temporal decision tree with minimum expected cost, and at the same time exploits entropy in order to keep the tree small.

Let us now see how FindSafeObs (also in Figure 12) computes the set of safe observations. Proposition 23 shows that the notion of safeness is tied to time labels rather than to individual observations. First of all FindSafeObs determines the range of valid time labels for the current set of examples (line 12); the lower bound $t_{low}$ is the lowest time label in Obs, and is stored in variable t_low, while the upper bound $t_{up}$ is given by the global deadline for Examples, and is stored in variable t_up.

Then the idea is to find the maximum safe label $t_{max}$ (variable t_max) which allows us to easily build the set of safe observations (line 21).

In order to accomplish this task the following steps have to be performed:

- Given the initial te-set $\mathbb{E}$, defined by Examples, Obs and ActModel, compute $\overline{\mathcal{X}}_{\mathbb{E},\mathbf{A}}$.

- For each time label $t$ in the range delimited by $t_{low}$ and $t_{up}$, consider the te-set $\mathbb{E}_t$ defined by Examples and by those observations with time label equal or greater than $t$. Then compute $\overline{\mathcal{X}}_{\mathbb{E}_t,\mathbf{A}}$.

- As soon as we find a time label $t$ with $\overline{\mathcal{X}}_{\mathbb{E}_t,\mathbf{A}} > \overline{\mathcal{X}}_{\mathbb{E},\mathbf{A}}$, we know that $t_{max}$ is the time label immediately preceding $t$.

Here the most critical operation (in terms of efficiency) is that of computing the expected cost of each $\mathbb{E}_t$, because this involves finding the quotient set $\mathbb{E}_t/\approx$. In fact, in order to obtain the quotient set, we need to repeatedly partition the te-set with respect to *all* observations available for it.

QuotientSet function (Figure 12) performs precisely this task. It takes in input the current time label tlabel, an initial partition (possibly made of a single block with the entire te-set) and the set of all observations, from which it will select valid ones.

First of all it partitions the input te-set with respect to observations with the current time label (lines 28–31). Then it executes iteratively the following operations:

- For each partition block it checks whether the deadline has moved further in time (lines 36–38).

- If so, it partitions again the block and stores the resulting sub-blocks for further examination (lines 39–41).

- If not, the block is part of the Final partition that will be returned (line 42).

In order to simplify the task, we introduce as a data type the *extended partition*, where each partition block is stored together with the highest time label used in building it. In this way we can easily check if the deadline for the block allows us to exploit more observations or not. Using extended partitions instead of standard ones we need to define a new function, ExtPartition, which works in the same way as the Partition function used in Figure 4, but also records with each block the highest time label used for it.

```
1    function ID3ChooseSafeObs (set Obs, te-set Examples, action model ActModel)
2    returns an observation o = ⟨sensor, tlabel⟩
3    begin
4        SafeObs ← FindSafeObs(Obs, Examples, ActModel);
5        o ← ID3ChooseTest(SafeObs, Examples);
6        return o;
7    end.
8    function FindSafeObs (set Obs, te-set Examples, action model ActModel)
9    returns a set of observations SafeObs
10   begin
11       cost ← ∑_{sit∈Examples} Act(sit);
12       t_up ← DI(Examples);   t_low ← min{t | ⟨s,t⟩ ∈ Obs};
13       t_max ← t_up;   part ← {⟨Examples, t_up⟩};
14       for every time label tx starting from t_up down to t_low do begin
15           part ← QuotientSet(part, Obs, tx);
16           newcost ← ExpectedCost(part);
17           if newcost < cost then begin
18               cost ← newcost;   t_max ← tx;
19           end;
20       end;
21       SafeObs ← {⟨s,t⟩ | t ≤ t_max};
22       return SafeObs;
23   end.
24   function QuotientSet (partition Initial, set Obs, time label tlabel)
25   returns a refined partition Final
26   begin
27       part ← Initial;
28       for all ⟨s,t⟩ with t = tlabel do begin
29           part ← ExtPartition(part, ⟨s,t⟩);
30           ObsCurr ← ObsCurr ∪ {⟨s,t⟩};
31       end;
32       Final ← ∅;
33       while part ≠ ∅ do
34           tmp_part ← ∅;
35           for each ⟨block, ty⟩ ∈ part do begin
36               newdl ← DI(block);
37               single ← {⟨block, ty⟩};
38               if newdl > ty then begin
39                   for each ⟨s,t⟩ with ty < t ≤ newdl do
40                       single ← ExtPartition(single, ⟨s,t⟩);
41                   tmp_part ← tmp_part ∪ single;
42               end else Final ← Final ∪ start;
43           end;
44           part ← tmp_part;
45       end;
46       return Final;
47   end.
```

Figure 12: ID3ChooseSafeObs is an implementation of ChooseObs yielding a tree with minimum expected cost

Notice that QuotientSet needs the whole set of observations (and not only valid ones) to properly compute the result; therefore when BuildTemporalTree calls ID3ChooseSafeObs it must pass as first argument UsefulObs instead of ValidObs.

FindSafeObs exploits QuotientSet to find all quotient sets for all $\mathbb{E}_t$, but does so using an efficient approach which we call *backward strategy*.

First of all, we can notice that the order in which observations are considered does not matter while building a quotient set. Moreover, if $t < t'$, $\mathbb{E}_t/\approx$ is a refinement of $\mathbb{E}_{t'}/\approx$; in other words we can obtain it from $\mathbb{E}_{t'}/\approx$ by simply refining the partition with additional observations.

Thus, *we can compute all quotient sets at the same time as we compute $\mathbb{E}/\approx$.*

FindSafeObs does exactly so: it computes all quotient sets and their expected cost starting from the last time label $t_{\mathbf{up}}$. Each quotient set is not built from scratch, but as a refinement of the previous one. This is the reason why QuotientSet (and ExtPartition as well) takes as first argument not a single set, but a partition. In this way, all quotient sets are computed with the same operations[8] needed to build $\mathbb{E}/\approx$. The next section analyzes in further detail complexity issues.

## 6.3 Complexity

In this section we aim at showing that the additional computations needed in building temporal decision trees do not lead to a higher asymptotical complexity than that we would get by using the standard **ID3** algorithm on the same set of examples (we discussed in section 5.1 the circumstances that could make such an approach feasible).

Essentially the difference between the two cases lies in the presence of FindSafeObs function. Wherever BuildTree calls ID3ChooseTest, BuildTemporalTree calls ID3ChooseSafeObs, which in turn calls both FindSafeObs and ID3ChooseTest.

Let us compare FindSafeObs and ID3ChooseTest, which are similar in many ways. The former repeatedly partitions the input te-set with respect to every available observation; then it computes entropy for each partition built in such a way. FindSafeObs builds just one partition by exploiting *all* available observations; in other words instead of using each observation to partition the initial te-set, it exploits it in order to refine an existing partition of the same te-set. Moreover, at each time label it computes the expected cost of the partition built so far. Essentially, if we denote by $N_S$ the number of sensors and with $N_T$ the number of time labels in the initial partition, we have roughly the following comparison:

- $N_S \times N_T$ (number of observations) entropy computations for ID3ChooseTest vs. $N_T$ expected cost computations for FindSafeObs.

- $N_S \times N_T$ partitions of the initial te-set for ID3ChooseTest vs. $N_S \times N_T$ refinements of existing partitions of the initial te-set for FindSafeObs.

Entropy and expected cost can be computed with roughly the same effort: both require retrieving some information for each element of each partition block, and to combine this information in some quite straightforward way. The complexity of this task depends only on the overall number of elements, and not on how they are distributed between different block of the partition. So even if expected cost is computed most of the times on finer partitions than entropy, the only thing that matters is that both are partitions of the same set, thus involving the same elements.

---

8. There is a slight overhead due to the need to find which observations should be used at each step.

Now let us examine the problem of creating a partition. This involves retrieving a value for each element of each block of the initial partition (which again depends only on the number of elements, and not on the number of blocks of the initial partition) and to properly assign the element to a new partition block depending on the original block and on the new value. The main difference in this case between starting with the whole te-set (creation) or with an initial partition (refinement), is the size of the new blocks that are being created, which are smaller in the second case. Dependent on how we implement the partition data type, this may make no difference, or may take less time for the refinement case. However, it never happens that refinement (corresponding to FindSafeObs function) requires more time than the creation (corresponding to ID3ChooseTest function) of a partition.

Therefore we can claim that FindSafeObs function has the same asymptotic complexity as function ID3ChooseTest. Thus also ID3ChooseSafeObs has the same asymptotic complexity as ID3ChooseTest, and we can conclude that BuildTree has the same asymptotic complexity as BuildTemporalTree.

## 6.4 An Example

In this section we shall show how our algorithm operates on the te-set in Figure 1 with respect to the action model in Figure 7.

Let us summarize the information the algorithm receives. Eight fault situations are involved; moreover we can exploit three sensors, each of which can show five different qualitative values: **h** - high, **n** - normal, **l** - low, **v** - very low, **z** - zero. Time labels correspond to natural numbers ranging from 0 to 7, and we assume they correspond to times measured by an internal clock which is started at the time of fault detection. There are four basic recovery actions $a, b, c, d$, such that $d \prec b \prec a$ and $d \prec c \prec a$. The set of compound recovery actions is thus $\mathbb{A} = \{\{a\}, \{b\}, \{c\}, \{b, c\}, \{d\}\}$; the ordering relation is pictured in 7, together with action costs.

BuildTemporalTree is first called on the whole te-set. None of the two terminating conditions is met (notice however that there are two observations that are not useful, since they do not discriminate: $\langle s_3, 0 \rangle$ and $\langle s_3, 1 \rangle$). Then the main function calls ID3ChooseSafeObs and consequently FindSafeObs. Since the global deadline is 2 we must check time labels 0, 1 and 2, starting from the last one.

Exploiting only observations with time label 2 we obtain the following partition:

$$\{\{\mathbf{sit}_1, \mathbf{sit}_6\}, \{\mathbf{sit}_2, \mathbf{sit}_5, \mathbf{sit}_7, \mathbf{sit}_8\}, \{\mathbf{sit}_3\}, \{\mathbf{sit}_4\}\}$$

However in order to find the expected cost we still have to check if for some partition block the deadline has changed; this happens for $\{\mathbf{sit}_1, \mathbf{sit}_6\}$ as well as for $\{\mathbf{sit}_3\}$ and $\{\mathbf{sit}_4\}$. For the last two blocks it does not change anything - they already contain only one element. As to the first block, the deadline is now 5 and thus it is possible to further split the partition. Therefore we obtain that the partition for time label 2 is:

$$\mathbb{P}_{t=2} = \{\{\mathbf{sit}_1\}, \{\mathbf{sit}_6\}, \{\mathbf{sit}_2, \mathbf{sit}_5, \mathbf{sit}_7, \mathbf{sit}_8\}, \{\mathbf{sit}_3\}, \{\mathbf{sit}_4\}\}$$

After finding the partition, the algorithm computes the expected cost, which turns out to be:

$$
\begin{aligned}
\mathbb{X}_{\mathbb{E},t=2} &= \chi(\textbf{Act}(\textbf{sit}_1)) \cdot \frac{1}{8} + \chi(\textbf{Act}(\textbf{sit}_6)) \cdot \frac{1}{8} + \chi(\textbf{Act}(\{\textbf{sit}_2, \textbf{sit}_5, \textbf{sit}_7, \textbf{sit}_8\})) \cdot \frac{1}{2} \\
&+ \chi(\textbf{Act}(\textbf{sit}_3)) \cdot \frac{1}{8} + \chi(\textbf{Act}(\textbf{sit}_4)) \cdot \frac{1}{8} \\
&= 100 \cdot \frac{1}{8} + 10 \cdot \frac{1}{8} + 70 \cdot \frac{1}{2} + 20 \cdot \frac{1}{8} + 50 \cdot \frac{1}{8} \\
&= 57.5
\end{aligned}
$$

Then the algorithm moves to time label 1; starting from $\mathbb{P}_{t=2}$ it adds observations with time label 1, obtaining a new partition:

$$
\mathbb{P}_{t=1} = \{\{\textbf{sit}_1\}, \{\textbf{sit}_6\}, \{\textbf{sit}_2, \textbf{sit}_5\}, \{\textbf{sit}_7\}, \{\textbf{sit}_8\}, \{\textbf{sit}_3\}, \{\textbf{sit}_4\}\}
$$

Deadlines move for $\{\textbf{sit}_7\}$ and $\{\textbf{sit}_8\}$, but since these are singletons the new observations cannot further split the partition. The expected cost is now:

$$
\begin{aligned}
\mathbb{X}_{\mathbb{E},t=1} &= \chi(\textbf{Act}(\textbf{sit}_1)) \cdot \frac{1}{8} + \chi(\textbf{Act}(\textbf{sit}_6)) \cdot \frac{1}{8} + \chi(\textbf{Act}(\{\textbf{sit}_2, \textbf{sit}_5\})) \cdot \frac{1}{4} + \chi(\textbf{Act}(\textbf{sit}_7)) \cdot \frac{1}{8} \\
&+ \chi(\textbf{Act}(\textbf{sit}_8)) \cdot \frac{1}{8} + \chi(\textbf{Act}(\textbf{sit}_3)) \cdot \frac{1}{8} + \chi(\textbf{Act}(\textbf{sit}_4)) \cdot \frac{1}{8} \\
&= 100 \cdot \frac{1}{8} + 10 \cdot \frac{1}{8} + 70 \cdot \frac{1}{4} + 20 \cdot \frac{1}{8} + + 50 \cdot \frac{1}{8} + 20 \cdot \frac{1}{8} + 50 \cdot \frac{1}{8} \\
&= 48.75
\end{aligned}
$$

Since $\mathbb{X}_{\mathbb{E},t=1} < \mathbb{X}_{\mathbb{E},t=2}$ we can conclude that observations with time label 2 are not safe. We now move to time label 0, and we immediately realize that the new observations do not change the partition. Thus $\mathbb{X}_{\mathbb{E},t=0} = \mathbb{X}_{\mathbb{E},t=1}$, and safe observations are those with time label equal either to 0 or to 1.

The algorithm now calls function ID3ChooseTest which selects the observation with minimum entropy. Figure 13 shows the entropies of the different observations at this stage, from which we deduce that the best choice is $\langle s_2, 1 \rangle$.

| Entropies for $\{\textbf{sit}_1, \textbf{sit}_2, \textbf{sit}_3, \textbf{sit}_4, \textbf{sit}_5, \textbf{sit}_6, \textbf{sit}_7, \textbf{sit}_8\}$ | | | |
|---|---|---|---|
| $\langle s_1, 0 \rangle$ | 1.5 | $\langle s_1, 1 \rangle$ | 1.5 |
| $\langle s_2, 0 \rangle$ | 1.451 | $\langle s_2, 1 \rangle$ | 0.844 |

Figure 13: Entropies for safe observations at the initial call

Figure 15.(a) shows the tree at this point; function BuildTemporalTree recursively invokes itself four times, yielding:

- a call $c_1$ on $\mathbb{E}_{\textbf{v}} = \{\textbf{sit}_1, \textbf{sit}_6\}$;
- a call $c_2$ on $\mathbb{E}_{\textbf{n}} = \{\textbf{sit}_2, \textbf{sit}_5\}$;
- a call $c_3$ on $\mathbb{E}_{\textbf{l}} = \{\textbf{sit}_3, \textbf{sit}_4, \textbf{sit}_7\}$;
- a call $c_4$ on $\mathbb{E}_{\textbf{h}} = \{\textbf{sit}_8\}$.

Let us focus on call $c_1$: again, none of the terminating conditions is met, therefore the algorithm invokes ID3ChooseSafeObs and thus FindSafeObs. Notice however that on this subset only a few observations are in UsefulObs: $\langle s_1, 3 \rangle$, $\langle s_2, 3 \rangle$, $\langle s_2, 4 \rangle$, $\langle s_2, 5 \rangle$ and $\langle s_3, 5 \rangle$. The global deadline is 5.

First we find the partition (and the expected cost) for time label 5:

$$\mathbb{P}_{t=5} = \{\{\mathbf{sit}_1\}, \{\mathbf{sit}_6\}\}$$
$$\mathbb{X}_{\mathbb{E}_\mathbf{v}, t=5} = \chi(\mathbf{Act}(\mathbf{sit}_1)) \cdot \frac{1}{2} + \chi(\mathbf{Act}(\mathbf{sit}_6)) \cdot \frac{1}{2} = 100 \cdot \frac{1}{2} + 10 \cdot \frac{1}{2} = 55$$

No additional observations can split further this partition and lower the cost; therefore we find that all valid observations are also safe. It is moreover obvious that all these observations have the same entropy, which is 0. Therefore the algorithm can non-deterministically select one of them; a reasonable criterion would be to select any of the earliest ones, for example $\langle s_1, 3 \rangle$.

Since the initial te-set for call $c_1$ is now split in two, there are two more recursive calls. However we can notice that if BuildTemporalTree is called on a te-set with a single element, the first terminating condition is trivially met (all the fault situations have the same recovery action). The function simply returns a tree leaf with the name of the proper recovery action. Figure 15.(b) shows the tree after $c_1$ has been completed.

Now let us examine $c_2$: the algorithm eliminates non-discriminating observations, and finds out that the set of useful observations is empty. Thus it builds a leaf with recovery action $\{b, c\}$. Let us pass to call $c_3$. In this case none of the terminating conditions is met:

| Entropies for $\{\mathbf{sit}_3, \mathbf{sit}_4, \mathbf{sit}_7\}$ | | | | | |
|---|---|---|---|---|---|
| $\langle s_1, 1 \rangle$ | 0.667 | $\langle s_1, 2 \rangle$ | 0.667 | $\langle s_1, 3 \rangle$ | 0.667 |
| $\langle s_2, 2 \rangle$ | 0.667 | $\langle s_2, 3 \rangle$ | 0.667 | $\langle s_2, 4 \rangle$ | 0.667 |
| $\langle s_2, 5 \rangle$ | 0 | $\langle s_3, 1 \rangle$ | 0 | $\langle s_3, 2 \rangle$ | 0 |
| $\langle s_3, 3 \rangle$ | 0 | $\langle s_3, 4 \rangle$ | 0 | | |

Figure 14: Entropies for safe observations at call $c_3$

the algorithm must then look for safe observations. The global deadline is 5, so we start examining time label 5, and we find:

$$\mathbb{P}_{t=5} = \{\{\mathbf{sit}_3\}, \{\mathbf{sit}_4\}, \{\mathbf{sit}_7\}\}$$
$$\mathbb{X}_{\mathbb{E}_\mathbf{t}, t=5} = \chi(\mathbf{Act}(\mathbf{sit}_3)) \cdot \frac{1}{3} + \chi(\mathbf{Act}(\mathbf{sit}_4)) \cdot \frac{1}{3} + \chi(\mathbf{Act}(\mathbf{sit}_7)) \cdot \frac{1}{3}$$
$$= 20 \cdot \frac{1}{3} + 50 \cdot \frac{1}{3} + 20 \cdot \frac{1}{3} = 30$$

Much as happened for $c_1$, no additional observation can further split the partition, so we can conclude that all valid observations are also safe. Figure 14 shows entropy for all valid observations; the earliest one with minimum entropy is $\langle s_3, 2 \rangle$ and this the algorithm selects. The two recursive sub-calls that are generated immediately terminate: $\{\mathbf{sit}_4\}$ is a singleton, and in $\{\mathbf{sit}_3, \mathbf{sit}_7\}$ both fault situations correspond to the same recovery action.

The last recursive call, $c_4$, has in input a singleton and thus immediately terminates. The final decision tree $\mathbf{T}$ is pictured in Figure 15.(c).

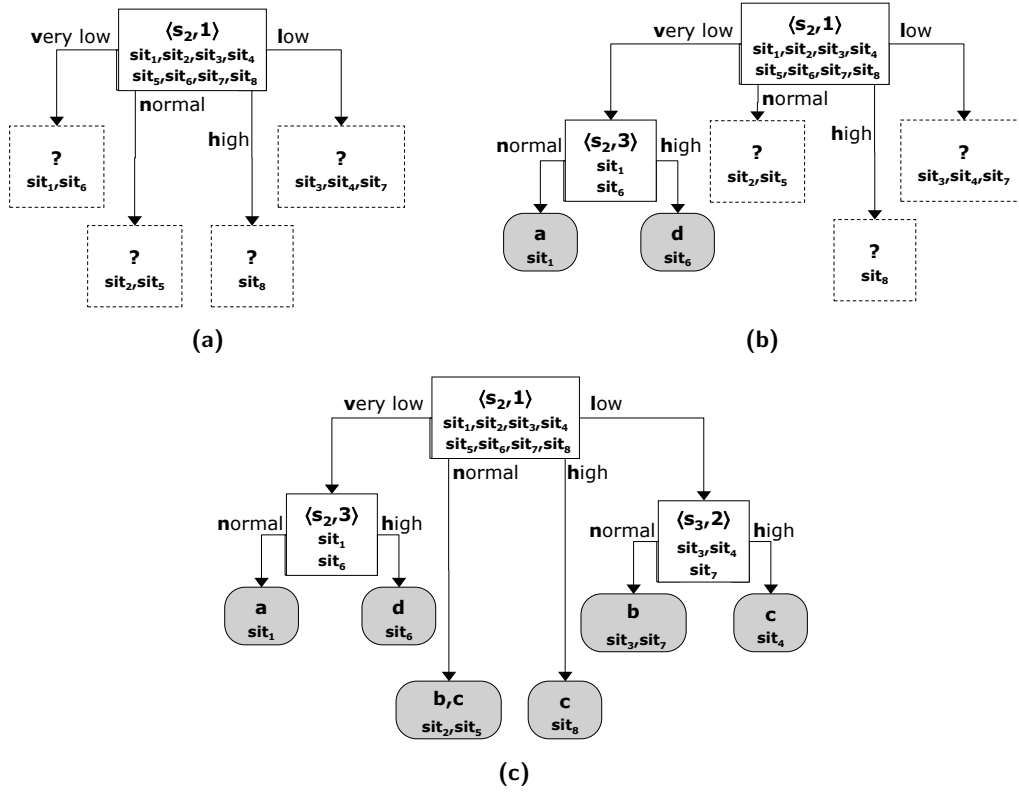

Figure 15: The output tree at different stages. **(c)** shows the final tree.

Let us check if the expected cost of **T** is really equal to the expected cost for $\mathbb{E}$. Figure 16 shows for each tree leaf $l$ the corresponding set of fault situations **examples**$(l)$, its probability $P(\textbf{examples}(l); \mathbb{E})$ and its cost $\chi(\mathcal{L}(l))$. Leaves are numbered from 1 to 6 going left to right in Figure 15.(c).

| leaf | examples | P | $\chi$ |
|------|----------|---|--------|
| $l_1$ | **sit**$_1$ | 1/8 | 100 |
| $l_2$ | **sit**$_6$ | 1/8 | 10 |
| $l_3$ | **sit**$_2$, **sit**$_5$ | 1/4 | 70 |
| $l_4$ | **sit**$_8$ | 1/8 | 50 |
| $l_5$ | **sit**$_3$, **sit**$_7$ | 1/4 | 20 |
| $l_6$ | **sit**$_4$ | 1/8 | 50 |

Figure 16: Fault situations, probabilities and costs for the leaves of the tree in Figure 15.(c)

502

Thus the expected cost of the tree is:

$$
\begin{aligned}
\mathbb{X}_{\mathbb{E},\mathbf{A}}(\mathbf{T}) &= \sum_{i=1}^{6} \chi(\mathcal{L}(l_i)) \cdot P(\mathbf{examples}(l); \mathbb{E}) \\
&= \frac{1}{8} \cdot 100 + \frac{1}{8} \cdot 10 + \frac{1}{4} \cdot 70 + \frac{1}{8} \cdot 50 + \frac{1}{4} \cdot 20 + \frac{1}{8} \cdot 50 \\
&= 48.75
\end{aligned}
$$

If we look back at example 8 we see that $\overline{\mathbb{X}}_{\mathbb{E},\mathbf{A}} = 48.75$, thus $\mathbf{T}$ has the minimum possible expected cost. Moreover, we can compare $\mathbf{T}$ with the tree $\mathbf{T}_1$ of example 5. Also $\mathbf{T}_1$ has the minimum possible expected cost, but $\mathbf{T}$ is more compact.

## 7. Conclusions

In this paper we introduced a new notion of diagnostic decision tree that takes into account temporal information on the observations and temporal constraints on the recovery actions to be performed. In this way we can take advantage of the discriminatory power that is available in the model of a dynamic system. We presented an algorithm that generates temporal diagnostic decision trees from a set of examples, discussing also how an optimal tree can be generated.

The automatic compilation of decision trees seems to be a promising approach for reconciling the advantages of model-based reasoning and the constraints imposed by the on-board hardware and software environment. It is worth noting that this is not only true in the automotive domain and indeed the idea of compiling diagnostic rules from a model has been investigated also in other approaches (see e.g., Darwiche, 1999; Dvorak & Kuipers, 1989). Darwiche (1999), in particular, discusses how rules can be generated for those platforms where constrained resources do not allow a direct use of a model-based diagnostic system.

What is new in our approach is the possibility of compiling also information concerning the system temporal behaviour, obtaining in this way more accurate decision procedures.

To the best of our knowledge, temporal decision trees are a new notion in the diagnostic literature. However, there are works in other fields that have some relation to ours, since they are aimed at learnig rules or associations that take into account time.

Geurts and Whenkel (1998) propose a notion of temporal tree to be used for early prediction of faults. This topic is closely related to diagnosis, albeit different in some ways: the idea is that the device under examination has not failed yet, but by observing its behaviour is possible to predict that a fault is about to occur. Geurts and Wehenkel propose to learn the relation between observed behavioural patterns and consequent failures by inducing a *temporal tree.*

The notion of temporal tree introduced by Geurts and Wehenkel is different than our temporal decision trees, reflecting the different purpose it has been introduced for. Rather than sensor readings, it consider a more general notion of test, and the tree does not specifies the time to wait before performing the tests, but rather the agent running the tree is supposed to wait until one of the tests associated to a tree node becomes true.

Also the notion of optimality is quite different: in the situation described by Geurts and Wehenkel the size of the resulting tree is not a concern. The tree-building algorithms aims then at minimizing the time at which the final decision is taken. In our algorithm, size is the primary concern, while from the point of view of time it suffices that diagnosis is carried out within certain deadlines. From the point of view of time alone, the approach by Geurts and Wehenkel is probably more general than ours; the problem of considering the trade-off between diagnostic capability and time needed for diagnosis is one of the major extensions we are considering for future work on this topic (see below).

Finally, the algorithm proposed by Geurts and Wehenkel works in a quite different way than ours: it first builds the tree greedily, using an evaluation function that weighs discriminability power agains time needed to reach a result, and selecting at each step the texts that optimizes such function. Then it prunes the tree in order to avoid overfitting. On the other hand, our approach aim at optimizing the tree from the point of view of cost, and at the same time tries to keep the tree small with the entropy heuristic. We think that, since optimization can be carried out at no additional cost[9] with respect to the minimization of entropy, our approach can obtain better results, at least in those cases where one can define a notion of deadline.

The process of learning association rules involving time has also been studied in other areas, such as machine learning (see for example Bischof & Caelli, 2001, where the authors propose a technique to learn movements) and data mining. While the specific diagnostic tailoring of our approach makes it difficult to compare it with more generic learning algorithms, the connections with data mining may be stronger. Our proposal in fact aims essentially at extracting from series of observations those patterns in time that allow to correctly diagnose a fault: this process can be regarded as a form of temporal classification. A preliminary investigation of papers in this area (see Antunes & Oliveira, 2001 for an overview) seems to suggest that, whereas the analysis of temporal sequences of data has received much interest in the last years, not much work has been done in the direction of data classification, where temporal decision trees could be exploited.

This suggests an interesting development for our work, in particular as concerns its applicability in other areas. However, we believe that the algorithm we presented needs to be extended in order to be exploited in other contexts. In particular we are investigating the following extensions:

- Deadlines could be turned from *hard* to *soft*. Soft deadlines do not have to be met, but rather define a cost associated to not meeting them. Thus not meeting a deadline becomes an option that can be taken into account when it is less expensive than performing a recovery action when the diagnosis is not complete. One could even define a cost that increases as the time passes from the expiration of the deadline. Such an extension would allow to model also the trade-off between discriminability power and time needed by the decision process, which we believe is the key to making our work applicable in other areas.

- Actions could be assumed to have a different cost depending on the fault situation; for example the action associated to a fault could become dangerous and thus extremely expensive if performed in presence of another fault.

---

9. From the point of view of asymptotical complexity.

On the long term, future work on this topic will be aimed at widening its areas of applicability, and investigating in deeper details its connections with other fields, such as fault prevention and data mining.

## 8. Acknowledgements

This work was partially supported by the EU under the grant GRD-1999-0058, project IDD (Integrated Diagnosis and Design), whose partners are: Centro Ricerche Fiat, Daimler-Chrysler, Magneti Marelli, OCC'M, PSA, Renault, Technische Universität München, Université Paris XIII, Università di Torino.

## Appendix A. Proofs

This section contains the proofs of all propositions, lemmas and theorems in the paper.

**Proposition 12.** *Let* $\mathbf{T} = \langle r, N, E, \mathcal{L}, \mathcal{T} \rangle$ *be a temporal decision tree compatible with a te-set* $\mathbb{E}$. *Let* $l_1, \ldots, l_f \in N$ *denote the leaves of* $T$. *Then* **examples**$(l_1), \ldots,$ **examples**$(l_f)$ *is a partition of* $\mathbb{E}$.

**Proof.** Follows immediately by the definition of **examples** (10), by noticing that if $n_1, \ldots, n_k$ are the children of $p$ then $\{$**examples**$(n_1), \ldots,$ **examples**$(n_k)\}$ is a partition of **examples**$(p)$. ◇

**Proposition 15** *Let* $\mathbf{T} = \langle r, N, E, \mathcal{L}, \mathcal{T} \rangle$ *denote a temporal decision tree, and let* $l_1, \ldots, l_u$ *be its leaves. Then*

$$\mathcal{X}_{\mathbb{E}, \mathbf{A}}(\mathbf{T}) = \sum_{i=1}^{u} \chi(\mathcal{L}(l_i)) \cdot P(\textbf{examples}(l); \mathbb{E})$$

**Proof.** By induction on the depth of $\mathbf{T}$. If $\mathbf{T}$ has depth 0 then it consists of a single leaf $l$ and 10 holds trivially since **examples**$(l)$ must be equal to $\mathbb{E}$ and $P(\mathbb{E}; \mathbb{E}) = 1$.

If $\mathbf{T}$ has depth $> 0$ then let $\mathbf{T}_1, \ldots, \mathbf{T}_k$ denote its direct subtrees and $c_1, \ldots, c_k$ denote their roots. We can regard each $\mathbf{T}_i$ as an autonomous temporal decision tree compatible with te-set $\mathbb{E}_i = $ **examples**$(c_i)$. By induction hypothesis we have that:

(13)
$$\mathcal{X}_{\mathbb{E}_i, \mathbf{A}}(\mathbf{T}_i) = \sum_{l \text{ leaf of } \mathbf{T}_i} \chi(\mathcal{L}(l)) \cdot P(\textbf{examples}(l); \mathbb{E}_i) = \sum_{l \text{ leaf of } \mathbf{T}_i} \chi(\mathcal{L}(l)) \cdot \frac{P(\textbf{examples}(l); \mathbb{E})}{P(\mathbb{E}_i; \mathbb{E})}.$$

Moreover by definition of expected cost:

(14) $$\mathcal{X}_{\mathbb{E}, \mathbf{A}}(\mathbf{T}) = \sum_{i=1}^{k} P(\mathcal{L}(r) \to \mathcal{L}((r, c_i))) \cdot \mathcal{X}_{\mathbb{E}_i, \mathbf{A}}(\mathbf{T}_i) \quad \text{with}$$

$$P(\mathcal{L}(r) \to \mathcal{L}((r, c_i))) = P(\textbf{examples}(c_i); \textbf{examples}(r)) = P(\mathbb{E}_i; \mathbb{E}).$$

From (13) and (14) we thus obtain:

$$(15) \quad \mathcal{X}_{\mathbb{E},\mathbf{A}}(\mathbf{T}) = \sum_{i=1}^{k} \sum_{l \text{ leaf of } \mathbf{T}_i} P(\mathbb{E}_i; \mathbb{E}) \cdot \chi(\mathcal{L}(l)) \cdot \frac{P(\mathbf{examples}(l); \mathbb{E})}{P(\mathbb{E}_i; \mathbb{E})}$$

$$= \sum_{i=1}^{k} \sum_{l \text{ leaf of } \mathbf{T}_i} \chi(\mathcal{L}(l)) \cdot P(\mathbf{examples}(l); \mathbb{E})$$

Since the leaves of $\mathbf{T}$ are all and only the leaves of $\mathbf{T}_1, \ldots, \mathbf{T}_k$, (15) is equivalent to the thesis. $\diamond$

**Proposition 16** *Let* $\mathbf{T} = \langle r_{\mathbf{T}}, N_{\mathbf{T}}, E_{\mathbf{T}}, \mathcal{L}_{\mathbf{T}}, \mathcal{T}_{\mathbf{T}} \rangle$, $\mathbf{U} = \langle r_{\mathbf{U}}, N_{\mathbf{U}}, E_{\mathbf{U}}, \mathcal{L}_{\mathbf{U}}, \mathcal{T}_{\mathbf{U}} \rangle$ *be two temporal decision trees compatible with the same te-set* $\mathbb{E}$ *and the same actions model* $\mathbf{A}$. *If* $\mathbf{T}$ *is more discriminating than* $\mathbf{U}$ *then* $\mathcal{X}_{\mathbb{E},\mathbf{A}}(\mathbf{T}) < \mathcal{X}_{\mathbb{E},\mathbf{A}}(\mathbf{U})$.

**Proof.** Rewriting equation (10) we obtain:

$$(16) \quad \mathcal{X}_{\mathbb{E},\mathbf{A}}(\mathbf{T}) = \sum_{i=1}^{r} \chi(\mathcal{L}_{\mathbf{T}}(l_i)) \cdot P(\mathbf{examples}(l); \mathbb{E}) = \sum_{\mathbf{sit} \in \mathbb{E}} \chi(\mathcal{L}_{\mathbf{T}}(\mathbf{leaf_T}(\mathbf{sit}))) P(\mathbf{sit}; \mathbb{E})$$

$$(17) \quad \mathcal{X}_{\mathbb{E},\mathbf{A}}(\mathbf{U}) = \sum_{\mathbf{sit} \in \mathbb{E}} \chi(\mathcal{L}_{\mathbf{U}}(\mathbf{leaf_U}(\mathbf{sit}))) P(\mathbf{sit}; \mathbb{E}).$$

Since $\mathbf{T}$ is more discriminating than $\mathbf{U}$, we have that for all $\mathbf{sit} \in \mathbb{E}$:

$$\mathcal{L}_{\mathbf{T}}(\mathbf{leaf_T}(\mathbf{sit})) \prec \mathcal{L}_{\mathbf{U}}(\mathbf{leaf_U}(\mathbf{sit})) \quad \text{or} \quad \mathcal{L}_{\mathbf{T}}(\mathbf{leaf_T}(\mathbf{sit})) = \mathcal{L}_{\mathbf{U}}(\mathbf{leaf_U}(\mathbf{sit}))$$

with at least one $\mathbf{sit}$ satisfying the first relation. By definition of $\chi$ it follows that for all $\mathbf{sit}$:

$$\chi(\mathcal{L}_{\mathbf{T}}(\mathbf{leaf_T}(\mathbf{sit}))) \leq \chi(\mathcal{L}_{\mathbf{U}}(\mathbf{leaf_U}(\mathbf{sit})))$$

and for at least one $\mathbf{sit}$:

$$\chi(\mathcal{L}_{\mathbf{T}}(\mathbf{leaf_T}(\mathbf{sit}))) = \chi(\mathcal{L}_{\mathbf{U}}(\mathbf{leaf_U}(\mathbf{sit})))$$

Therefore if we compare the individual elements of the two sums in (16) and (17) we observe there exists at least one $\mathbf{sit}$ for which:

$$\chi(\mathcal{L}_{\mathbf{T}}(\mathbf{leaf_T}(\mathbf{sit}))) P(\mathbf{sit}; \mathbb{E}) < \chi(\mathcal{L}_{\mathbf{U}}(\mathbf{leaf_U}(\mathbf{sit}))) P(\mathbf{sit}; \mathbb{E})$$

and for all other $\mathbf{sit}$

$$\chi(\mathcal{L}_{\mathbf{T}}(\mathbf{leaf_T}(\mathbf{sit}))) P(\mathbf{sit}; \mathbb{E}) \leq \chi(\mathcal{L}_{\mathbf{U}}(\mathbf{leaf_U}(\mathbf{sit}))) P(\mathbf{sit}; \mathbb{E})$$

which concludes the proof. $\diamond$

**Theorem 20** *Let* $\mathbb{E}$ *be a te-set with actions model* $\mathbf{A}$. *We have that:*

(i) *There exists a decision tree* $\overline{\mathbf{T}}$ *compatible with* $\mathbb{E}$ *such that* $\mathcal{X}_{\mathbb{E},\mathbf{A}}(\overline{\mathbf{T}}) = \overline{\mathcal{X}}_{\mathbb{E},\mathbf{A}}$.

(ii) *For every temporal decision tree* $\mathbf{T}$ *compatible with* $\mathbb{E}$, $\overline{\mathcal{X}}_{\mathbb{E},\mathbf{A}} \leq \mathcal{X}_{\mathbb{E},\mathbf{A}}(\mathbf{T})$. $\diamond$

In order to prove this theorem we introduce some lemmas.

**Lemma 26** *Let $\mathbf{T}$ be a temporal decision tree compatible with a te-set $\mathbb{E}$. Then $\mathbf{sit}_i \approx \mathbf{sit}_j$ implies $\mathbf{leaf_T}(\mathbf{sit}_i) = \mathbf{leaf_T}(\mathbf{sit}_j)$.*

**Proof.** We prove that $\mathbf{sit}_i \sim \mathbf{sit}_j$ implies $\mathbf{leaf_T}(\mathbf{sit}_i) = \mathbf{leaf_T}(\mathbf{sit}_j)$, from which the lemma easily follows. Let us suppose that $\mathbf{leaf_T}(\mathbf{sit}_i) \neq \mathbf{leaf_T}(\mathbf{sit}_j)$. This means that there is a common ancestor $n$ of the two leaves such that $\mathbf{sit}_i, \mathbf{sit}_j \in \mathbf{examples}(n)$ and $\mathbf{Val}(\mathbf{sit}_i, \langle \mathcal{L}(n), \mathcal{T}(n) \rangle) \neq \mathbf{Val}(\mathbf{sit}_j, \langle \mathcal{L}(n), \mathcal{T}(n) \rangle)$. Since $\mathbf{sit}_i \sim \mathbf{sit}_j$ this is possible only if $\mathcal{T}(n) > \min\{\mathbf{Dl}(\mathbf{sit}_i), \mathbf{Dl}(\mathbf{sit}_j)\}$. But since $\mathbf{T}$ is compatible with $\mathbb{E}$ it must hold that $\mathcal{T}(n) \leq \mathbf{Dl}(\mathbf{examples}(n)) \leq \min\{\mathbf{Dl}(\mathbf{sit}_i), \mathbf{Dl}(\mathbf{sit}_j)\}$, which contradicts the previous statement. ◇

**Lemma 27** *Let $\mathbb{E}$ be a te-set with sensors $s_1, \ldots, s_m$, time labels $t_1, \ldots, t_{\mathsf{last}}$ and actions model $\mathbf{A}$. There exists a temporal decision tree $\overline{\mathbf{T}} = \langle \overline{r}, \overline{N}, \overline{E}, \overline{\mathcal{L}}, \overline{\mathcal{T}} \rangle$ such that $\mathcal{X}_{\mathbb{E}, \mathbf{A}}(\overline{\mathbf{T}}) = \overline{\mathcal{X}}_{\mathbb{E}, \mathbf{A}}$.*

**Proof.** In order to prove the thesis we construct a tree $\overline{\mathbf{T}}$ with the same expected cost as the te-set.

Let us define a total order on observations in $\mathbb{E}$ as follows: $\langle s, t \rangle < \langle s', t' \rangle$ if either $t < t'$ or $t = t'$ and $s$ precedes $s'$ in a lexicographic ordering. Let us denote by $o_1, \ldots, o_{\mathsf{max}}$ the ordered sequence of observations thus obtained. We shall define $\overline{\mathbf{T}}$ level by level (starting from the root, at level 1) giving the value of $\mathcal{L}$ and $\mathcal{T}$ for nodes at level $h$.

$\overline{\mathbf{T}}$ has a maximum of $\mathsf{max} + 1$ levels, where $\mathsf{max}$ is the number of observations. New levels are added until all nodes in a level are leaves (which as we shall see happens at most at level $\mathsf{max} + 1$). Let thus $n$ be a node at level $h$, and let $\langle s_{i_h}, t_{i_h} \rangle = o_h$ if $h \leq \mathsf{max}$. We have:

$$n \text{ is } \begin{cases} \text{a } \textit{leaf} & \text{if } h = \mathsf{max} + 1, \text{ or } \mathbf{Dl}(\mathbf{examples}(n)) < t_{i_h}; \\ \text{an } \textit{internal node} & \text{otherwise.} \end{cases}$$

$$\mathcal{L}(n) = \begin{cases} \mathbf{merge}(\{\mathbf{Act}(\mathbf{sit} \mid \mathbf{sit} \in \mathbf{examples}(n)\}) & \text{if } n \text{ is a leaf;} \\ s_{i_h} & \text{if } n \text{ is an internal node.} \end{cases}$$

$$\mathcal{T}(n) = t_{i_h} \quad \text{if } n \text{ is an internal node.}$$

A decision-making agent running such a tree would essentially take into account *all* sensor measurement at *all* time labels until either there are no more available observations or it must perform a recovery action because a deadline is about to expire.

Now we need to show that $\mathcal{X}_{\mathbb{E}, \mathbf{A}}(\overline{\mathbf{T}}) = \overline{\mathcal{X}}_{\mathbb{E}, \mathbf{A}}$.

Let $l_1, \ldots, l_u$ denote the leaves of $\overline{\mathbf{T}}$. We shall first of all prove that $\mathbf{sit}_i \approx \mathbf{sit}_j$ if and only if $\mathbf{leaf}_{\overline{\mathbf{T}}}(\mathbf{sit}_i) = \mathbf{leaf}_{\overline{\mathbf{T}}}(\mathbf{sit}_j)$, or equivalently that

$$\{\mathbf{examples}(l_1), \ldots, \mathbf{examples}(l_u)\} = \mathbb{E}/\approx.$$

This, together with equations (10) and (11) yields the thesis.

We already know from lemma 26 that $\mathbf{sit}_i \approx \mathbf{sit}_j$ implies $\mathbf{leaf}_{\overline{\mathbf{T}}}(\mathbf{sit}_i) = \mathbf{leaf}_{\overline{\mathbf{T}}}(\mathbf{sit}_j)$; we need to show that the opposite is also true. Let us thus assume that $\mathbf{sit}_i, \mathbf{sit}_j \in \mathbf{examples}(l)$ for some $l \in \{l_1, \ldots, l_u\}$. Let $r = n_1, n_2, \ldots, n_H, n_{H+1} = l$ be the path from the root to $l$. We know from the definition of $\overline{\mathbf{T}}$ that for all $h \leq H$, $\langle \mathcal{L}(n_h), \mathcal{T}(n_h) \rangle = o_h$, and that $o_1, \ldots, o_H$ are *all* observations $\langle s, t \rangle$ of $\mathbb{E}$ such that $t \leq \mathbf{DI}(\mathbf{examples}(l))$. Moreover since $\mathbf{sit}_i, \mathbf{sit}_j \in \mathbf{examples}(l)$ we have that for all $h = 1, \ldots, H$, $\mathbf{Val}(\mathbf{sit}_i, o_h) = \mathbf{Val}(\mathbf{sit}_j, o_h)$.

Now there are two possibilities: either $\mathbf{DI}(\mathbf{examples}(l)) = \min\{\mathbf{DI}(\mathbf{sit}_i), \mathbf{DI}(\mathbf{sit}_j)\}$, or $\mathbf{DI}(\mathbf{examples}(l)) < \min\{\mathbf{DI}(\mathbf{sit}_i), \mathbf{DI}(\mathbf{sit}_j)\}$.

In the first case we immediately obtain that $\mathbf{sit}_i \sim \mathbf{sit}_j$ and thus $\mathbf{sit}_i \approx \mathbf{sit}_j$.

In the second case, there must be $\mathbf{sit}_k \in \mathbf{examples}(l)$ such that $\mathbf{DI}(\mathbf{examples}(l)) = \mathbf{DI}(\mathbf{sit}_k)$. Moreover, $\mathbf{DI}(\mathbf{sit}_k) = \min\{\mathbf{DI}(\mathbf{sit}_i), \mathbf{DI}(\mathbf{sit}_j)\} = \min\{\mathbf{DI}(\mathbf{sit}_j), \mathbf{DI}(\mathbf{sit}_k)\}$. Since all considerations above apply also to $\mathbf{sit}_k$ we thus have that $\mathbf{sit}_i \approx \mathbf{sit}_k$ and $\mathbf{sit}_k \approx \mathbf{sit}_j$; therefore by transitivity $\mathbf{sit}_i \approx \mathbf{sit}_j$. ◇

**Lemma 28** *Let* $\mathbf{T} = \langle r, N, E, \mathcal{L}, \mathcal{T} \rangle$ *be a decision tree compatible with a te-set* $\mathbb{E}$ *with actions model* $\mathbf{A}$. *Then* $\overline{\mathcal{X}}_{\mathbb{E}, \mathbf{A}} \leq \mathcal{X}_{\mathbb{E}, \mathbf{A}}(T)$.

**Proof.** Let $\overline{\mathbf{T}}$ be as defined in the proof of lemma 27. In order to prove the thesis it suffices to show that $\overline{\mathbf{T}}$ is either equally[10] or more discriminating than $\mathbf{T}$ (see proposition 16). Actually we shall show that given $\mathbf{sit} \in \mathbb{E}$ either $\overline{\mathcal{L}}(\mathbf{leaf}_{\overline{\mathbf{T}}}(\mathbf{sit})) \prec \mathcal{L}(\mathbf{leaf}_{\mathbf{T}}(\mathbf{sit}))$ or $\overline{\mathcal{L}}(\mathbf{leaf}_{\overline{\mathbf{T}}}(\mathbf{sit})) = \mathcal{L}(\mathbf{leaf}_{\mathbf{T}}(\mathbf{sit}))$.

We know that $\mathbf{examples}(\mathbf{leaf}_{\overline{\mathbf{T}}}(\mathbf{sit})) \subseteq \mathbf{examples}(\mathbf{leaf}_{\mathbf{T}}(\mathbf{sit}))$. In fact, let $\mathbf{sit}'$ be an element of $\mathbf{examples}(\mathbf{leaf}_{\overline{\mathbf{T}}}(\mathbf{sit}))$ different from $\mathbf{sit}$ itself: by construction of $\overline{\mathbf{T}}$ we have that $\mathbf{sit} \approx \mathbf{sit}'$, and by lemma 26 it follows that $\mathbf{leaf}_{\mathbf{T}}(\mathbf{sit}) = \mathbf{leaf}_{\mathbf{T}}(\mathbf{sit}')$.

Let:

$$\overline{A} = \{\mathbf{Act}(s) \mid s \in \mathbf{examples}(\mathbf{leaf}_{\overline{\mathbf{T}}}(\mathbf{sit}))\}, A = \{\mathbf{Act}(s) \mid s \in \mathbf{examples}(\mathbf{leaf}_{\mathbf{T}}(\mathbf{sit}))\}.$$

Since $\overline{A} \subseteq A$, by definition of $\mathbf{merge}$:

$$\mathbf{merge}(\overline{A}) \prec \mathbf{merge}(A) \quad \text{or} \quad \mathbf{merge}(\overline{A}) = \mathbf{merge}(A)$$

Thus having $\mathcal{L}(\mathbf{leaf}_{\overline{\mathbf{T}}}(\mathbf{sit})) = \mathbf{merge}(\overline{A})$ and $\mathcal{L}(\mathbf{leaf}_{\mathbf{T}}(\mathbf{sit})) = \mathbf{merge}(A)$ we obtain that either the action selected by $\overline{\mathbf{T}}$ is weaker than that selected by $\mathbf{T}$, or it is the same. ◇

Now we can prove theorem 20.

**Proof.** Point *(i)* is proved by lemma 27, while point *(ii)* corresponds to lemma 28. ◇

**Proposition 21** *Let us consider an execution of* BUILDTEMPORALTREE *starting with a main call* $c_0$. *The initial te-set, which we want to build a tree over, is* $\mathbb{E} = \mathbb{E}_{c_0}$ *with* $\mathbf{A} = [\![\mathtt{ActModel}]\!]_{c_0}$. *For any recursive call* $c$, *let us denote by* $\mathbb{E}_c^*$ *the te-set determined by* $[\![\mathtt{Examples}]\!]_c$ *and* $[\![\mathtt{Obs\_Update}]\!]_c$. *Then:*

**(1)** $\mathcal{X}_{\mathbb{E}, \mathbf{A}}([\![\mathbb{T}]\!]_{c_0}) \geq \overline{\mathcal{X}}_{\mathbb{E}, \mathbf{A}}$

**(2)** $\mathcal{X}_{\mathbb{E}, \mathbf{A}}([\![\mathbb{T}]\!]_{c_0}) = \overline{\mathcal{X}}_{\mathbb{E}, \mathbf{A}}$ *if and only if for every non terminal[11] recursive call* $c$ *generated by* $c_0$ *it holds that* $\overline{\mathcal{X}}_{\mathbb{E}_c, \mathbf{A}} = \overline{\mathcal{X}}_{\mathbb{E}_c^*, \mathbf{A}}$

---

10. Rather intuitively, two trees are equally discriminating if they associate to each fault situation the same recovery action.

11. We exclude terminal calls because they do not even compute $\mathtt{Obs\_Update}$.

**Proof.** We shall prove **(1)** and **(2)** for every recursive call $c$ (rather than only for $c_0$). The proof is by induction on the depth of the recursion starting from $c$.

**depth $= 0$.** Then $c$ is terminal, and we only have to prove that $\mathcal{X}_{\mathbb{E},\mathbf{A}}([\![\mathtt{T}]\!]_c) = \overline{\mathcal{X}}_{\mathbb{E},\mathbf{A}}$. There are two reasons why $c$ may be terminal: either *(i)* all fault situations in $[\![\mathtt{Examples}]\!]_c$ are associated with the same recovery action $A$, or *(ii)* $[\![\mathtt{ValidObs}]\!]_c = \varnothing$.

  (i) From definition 19 we have:

$$\overline{\mathcal{X}}_{\mathbb{E},\mathbf{A}} = \sum_{\eta \in \mathbb{E}_c/\approx} \chi(\mathbf{merge}(\{\mathbf{Act}(\mathbf{sit}) \mid \mathbf{sit} \in \eta\})) \cdot P(\eta; \mathbb{E}_c)$$

$$= \sum_{\eta \in \mathbb{E}_c/\approx} \chi(A) \cdot P(\eta; \mathbb{E}_c)$$

$$= \chi(A) \sum_{\eta \in \mathbb{E}_c/\approx} P(\eta; \mathbb{E}_c) = \chi(A).$$

  Since $[\![\mathtt{T}]\!]_c$ is made of a single leaf $l$ with $\mathcal{L}(l) = A$ we also have that $\mathcal{X}_{\mathbb{E}_c,\mathbf{A}}([\![\mathtt{T}]\!]_c) = \chi(A)$, which proves the thesis.

  (ii) If $[\![\mathtt{ValidObs}]\!]_c = \varnothing$ then for all $\langle s, t \rangle \in [\![\mathtt{Examples}]\!]_c$, $t > \mathbf{DI}([\![\mathtt{Examples}]\!]_c)$. Let $\mathbf{sit} \in \mathbb{E}_c$ be such that $\mathbf{DI}(\mathbf{sit}) = \mathbf{DI}([\![\mathtt{Examples}]\!]_c)$. Then by definition of indistinguishability for any $\mathbf{sit}' \in \mathbb{E}_c$ we have that $\mathbf{sit} \sim \mathbf{sit}'$. This proves that $\mathbb{E}_c/\approx$ is made of a single equivalence class which coincides with $\mathbb{E}_c$ itself. Thus $\overline{\mathcal{X}}_{\mathbb{E}_c,\mathbf{A}} = \chi(\mathbf{merge}(\{\mathbf{Act}(\mathbf{sit}) \mid \mathbf{sit} \in \mathbb{E}_c\}))$. Since $[\![\mathtt{T}]\!]_c$ is made of a single leaf $l$ with $\mathcal{L}(l) = \mathbf{merge}(\{\mathbf{Act}(\mathbf{sit}) \mid \mathbf{sit} \in \mathbb{E}_c\})$, it follows that $\mathcal{X}_{\mathbb{E}_c,\mathbf{A}}([\![\mathtt{T}]\!]_c) = \overline{\mathcal{X}}_{\mathbb{E}_c,\mathbf{A}}$.

**depth $> 0$.** Then $c$ is not terminal and ChooseObs selects an observation $o = \langle s, t \rangle$. Let $v_1, \ldots, v_k$ be the possible values for $o$: then $c$ has $k$ inner recursive calls to BuildTemporalTree, which we shall denote respectively by $c_1, \ldots, c_k$. We have that $\{\mathbb{E}_{c_1}, \ldots, \mathbb{E}_{c_k}\}$ is a partition of $\mathbb{E}_c^*$.

By definition of expected cost (14) we have that:

$$\mathcal{X}_{\mathbb{E}_c,\mathbf{A}}([\![\mathtt{T}]\!]_c) = \sum_{i=1}^{k} P(\mathbb{E}_c\!\mid_{o \to v_i}; \mathbb{E}_c) \cdot \mathcal{X}_{\mathbb{E}_c\!\mid_{o \to v_i},\mathbf{A}}([\![\mathtt{T}]\!]_{c_i})$$

$\mathbb{E}_c\!\mid_{o \to v_i}$ and $\mathbb{E}_{c_i}$ differ only in the set of observations, which is $[\![\mathtt{Obs}]\!]_c$ for the former and $[\![\mathtt{Obs\_Update}]\!]_c$ for the latter. However we have that:

- $P(\mathbb{E}_c\!\mid_{o \to v_i}; \mathbb{E}_c) = P(\mathbb{E}_{c_i}; \mathbb{E}_c)$ since probabilities depend only on the fault situations in a te-set, and not on the observations.
- $\mathcal{X}_{\mathbb{E}_c\!\mid_{o \to v_i},\mathbf{A}}([\![\mathtt{T}]\!]_{c_i}) = \mathcal{X}_{\mathbb{E}_{c_i},\mathbf{A}}([\![\mathtt{T}]\!]_{c_i})$: expected cost depends also on the observations, but $[\![\mathtt{T}]\!]_{c_i}$ by construction can contain as labels only those observations in $\mathbb{E}_{c_i}$.

Moreover, since also $\mathbb{E}_c$ and $\mathbb{E}_c^*$ differ only in the observations $P(\mathbb{E}_{c_i}; \mathbb{E}_c) = P(\mathbb{E}_{c_i}; \mathbb{E}_c^*)$. Therefore we can write:

$$\mathcal{X}_{\mathbb{E}_c,\mathbf{A}}([\![\mathtt{T}]\!]_c) = \sum_{i=1}^{k} P(\mathbb{E}_{c_i}; \mathbb{E}_c^*) \cdot \mathcal{X}_{\mathbb{E}_{c_i},\mathbf{A}}([\![\mathtt{T}]\!]_{c_i})$$

In order to prove **(1)**, we can apply the induction hypothesis $\mathcal{X}_{\mathbb{E}_{c_i}, \mathbf{A}}(\llbracket \mathbf{T} \rrbracket_{c_i}) \geq \overline{\mathcal{X}}_{\mathbb{E}_{c_i}, \mathbf{A}})$ and obtain:

$$(18) \qquad \mathcal{X}_{\mathbb{E}_c, \mathbf{A}}(\llbracket \mathbf{T} \rrbracket_c) \geq \sum_{i=1}^{k} P(\mathbb{E}_{c_i}; \mathbb{E}_c^*) \cdot \overline{\mathcal{X}}_{\mathbb{E}_{c_i}, \mathbf{A}}$$

Now let us work on the right-hand side expression in 18:

$$\begin{aligned}
\sum_{i=1}^{k} P(\mathbb{E}_{c_i}; \mathbb{E}_c^*) \cdot \overline{\mathcal{X}}_{\mathbb{E}_{c_i}, \mathbf{A}} &= \sum_{i=1}^{k} P(\mathbb{E}_{c_i}; \mathbb{E}_c^*) \sum_{\eta \in \mathbb{E}_{c_i}/\approx} \chi(\mathbf{Act}(\eta)) \cdot P(\eta; \mathbb{E}_{c_i}) \\
&= \sum_{i=1}^{k} \sum_{\eta \in \mathbb{E}_{c_i}/\approx} \chi(\mathbf{Act}(\eta)) \cdot P(\eta; \mathbb{E}_{c_i}) \cdot P(\mathbb{E}_{c_i}; \mathbb{E}_c^*) \\
&= \sum_{i=1}^{k} \sum_{\eta \in \mathbb{E}_{c_i}/\approx} \chi(\mathbf{Act}(\eta)) \cdot P(\eta; \mathbb{E}_c^*)
\end{aligned}$$

Notice however that $\{\mathbb{E}_{c_i}/\approx\}$ is a partition of $\mathbb{E}_c^*/\approx$; in other words each $\eta \in \mathbb{E}_c^*/\approx$ belongs to exactly one set $\mathbb{E}_{c_i}/\approx$. In fact, splitting examples according to the value of one observation cannot split a class of undistinguishable observations. Thus the above equality becomes:

$$\sum_{i=1}^{k} P(\mathbb{E}_{c_i}; \mathbb{E}_c^*) \cdot \overline{\mathcal{X}}_{\mathbb{E}_{c_i}, \mathbf{A}} = \sum_{\eta \in \mathbb{E}_c^*/\approx} \chi(\mathbf{Act}(\eta)) \cdot P(\eta; \mathbb{E}_c^*) = \overline{\mathcal{X}}_{\mathbb{E}_c^*, \mathbf{A}}$$

This, together with 18, yields:

$$(19) \qquad \mathcal{X}_{\mathbb{E}_c, \mathbf{A}}(\llbracket \mathbf{T} \rrbracket_c) \geq \overline{\mathcal{X}}_{\mathbb{E}_c^*, \mathbf{A}}$$

As mentioned above, the only difference between $\mathbb{E}_c^*$ and $\mathbb{E}_c$ is that the former has fewer observations. This implies that, if $\mathbf{sit} \approx \mathbf{sit}'$ in $\mathbb{E}_c$, then $\mathbf{sit} \approx \mathbf{sit}'$ in $\mathbb{E}_c^*$ as well. This means that $\mathbb{E}_c/\approx$ is a *sub-partition*[12] of $\mathbb{E}_c^*/\approx$ in the following sense: we can partition every $\theta \in \mathbb{E}_c^*/\approx$ in $\eta(\theta) = \{\eta_1', \ldots, \eta_{k_\theta}'\}$ such that for each $\eta_j'$ there exists exactly one $\eta_j \in \mathbb{E}_c/\approx$ containing exactly the same fault situations as $\eta_j'$. This yields:

$$\overline{\mathcal{X}}_{\mathbb{E}_c, \mathbf{A}} = \sum_{\eta \in \mathbb{E}_c/\approx} \chi(\mathbf{Act}(\eta)) \cdot P(\eta; \mathbb{E}_c)$$

Since each $\eta_j'$ has the same fault situations as the corresponding $\eta_j$, and necessarily for each $\eta_j$ there is a $\theta$ containing it, we have:

$$\overline{\mathcal{X}}_{\mathbb{E}_c, \mathbf{A}} = \sum_{\theta \in \mathbb{E}_c^*/\approx} \sum_{\eta' \in \eta(\theta)} \chi(\mathbf{Act}(\eta')) \cdot P(\eta; \mathbb{E}_c^*)$$

---

12. $\mathbb{E}_c/\approx$ is not a sub-partition of $\mathbb{E}_c^*/\approx$ in the ordinary sense because they do not have the same set of observations.

If $\eta' \in \eta(\theta)$ the fault situations in $\eta'$ are a subset of those in $\theta$; thus $\chi(\textbf{Act}(\eta')) \leq \chi(\textbf{Act}(\theta))$. Then we obtain:

$$\begin{aligned}
\overline{\mathcal{X}}_{\mathbb{E}_c, \mathbf{A}} &\leq \sum_{\theta \in \mathbb{E}_c^* / \approx} \sum_{\eta' \in \eta(\theta)} \chi(\textbf{Act}(\theta)) \cdot P(\eta'; \mathbb{E}_c^*) \\
&= \sum_{\theta \in \mathbb{E}_c^* / \approx} \chi(\textbf{Act}(\theta)) \cdot \sum_{\eta' \in \eta(\theta)} P(\eta'; \mathbb{E}_c^*) \\
&= \sum_{\theta \in \mathbb{E}_c^* / \approx} \chi(\textbf{Act}(\theta)) \cdot P(\theta; \mathbb{E}_c^*) = \overline{\mathcal{X}}_{\mathbb{E}_c, \mathbf{A}}
\end{aligned}$$

Together with equation 19, this proves **(1)**:

$$\mathcal{X}_{\mathbb{E}_c, \mathbf{A}}(\llbracket \texttt{T} \rrbracket_c) \geq \overline{\mathcal{X}}_{\mathbb{E}_c, \mathbf{A}}$$

Now let us prove **(2)**. The induction hypothesis changes 18, and thus 19, into equalities, thus yielding:

(20) $$\mathcal{X}_{\mathbb{E}_c, \mathbf{A}}(\llbracket \texttt{T} \rrbracket_c) = \overline{\mathcal{X}}_{\mathbb{E}_c^*, \mathbf{A}}$$

Since by hypothesis in (2) we have that $\overline{\mathcal{X}}_{\mathbb{E}_c^*, \mathbf{A}} = \overline{\mathcal{X}}_{\mathbb{E}_c, \mathbf{A}}$, we immediately obtain:

$$\mathcal{X}_{\mathbb{E}_c, \mathbf{A}}(\llbracket \texttt{T} \rrbracket_c) = \overline{\mathcal{X}}_{\mathbb{E}_c, \mathbf{A}}$$

which concludes the proof. ◇

**Proposition 22** *Let $c, d$ denote two independent calls to* Build­Temporal­Tree *with the same input arguments but with different implementations of* ChooseObs. *If* $\llbracket \texttt{tlabel} \rrbracket_c \leq \llbracket \texttt{tlabel} \rrbracket_d$ *then* $\overline{\mathcal{X}}_{\mathbb{E}_c^*, \mathbf{A}} \leq \overline{\mathcal{X}}_{\mathbb{E}_d^*, \mathbf{A}}$.

**Proof.** Follows immediately from $\llbracket \texttt{Obs\_Update} \rrbracket_c \subseteq \llbracket \texttt{Obs\_Update} \rrbracket_d$. ◇

**Proposition 23** *Let $c$ be a call to* Build­Temporal­Tree. *If* $\llbracket \texttt{tlabel} \rrbracket_c = t_{\textbf{min}_c} = \min\{t \mid \langle t, s \rangle \in \llbracket \texttt{ValidObs} \rrbracket_c\}$ *then* $\overline{\mathcal{X}}_{\mathbb{E}_c^*, \mathbf{A}} \leq \overline{\mathcal{X}}_{\mathbb{E}_c, \mathbf{A}}$.

**Proof.** In this case $\llbracket \texttt{Obs\_Update} \rrbracket_c = \llbracket \texttt{UsefulObs} \rrbracket_c$, thus the only removed observations are non discriminating ones. ◇

**Proposition 25** *For any call $c$ to* Build­Temporal­Tree *there exist a time label $t_{\textbf{max}_c}$ such that the safe time labels are all and only those $t$ with $t_{\textbf{min}_c} \leq t \leq t_{\textbf{max}_c}$, where $t_{\textbf{min}_c}$ is as in proposition 23.*

**Proof.** Straightforward. ◇